%% file: main.tex
\documentclass[10pt,twocolumn,letterpaper]{article}

\usepackage[pagenumbers]{cvpr} 

\input{preamble}
\definecolor{cvprblue}{rgb}{0.21,0.49,0.74}
\usepackage[pagebackref,breaklinks,colorlinks,allcolors=cvprblue]{hyperref}

\usepackage{nicematrix}
\usepackage{xcolor}
\usepackage{multirow}
\newcommand{\best}[1]{\textbf{#1}}
\newcommand{\seco}[1]{\underline{#1}}
\definecolor{Gray}{gray}{0.95}
\usepackage{colortbl}
\newcolumntype{a}{>{\columncolor{Gray}}c}

\usepackage{tcolorbox} %

\def\paperID{xxx} 
\def\confName{CVPR}
\def\confYear{2026}

\title{Wan-Weaver: Interleaved Multi-modal Generation via Decoupled Training}

\author{
Jinbo Xing${^{1,*}}$~~ 
Zeyinzi Jiang${^{1,*}}$~~ 
Yuxiang Tuo${^{1,*}}$~~
Chaojie Mao${^{1,*}}$~~
Xiaotang Gai${^1}$~~ 
Xi Chen${^1}$\\ 
Jingfeng Zhang${^1}$~~
Yulin Pan${^1}$~~ 
Zhen Han${^1}$~~
Jie Xiao${^1}$~~ 
Keyu Yan${^1}$~~ 
Chenwei Xie${^1}$\\ 
Chongyang Zhong${^1}$~~ 
Kai Zhu${^1}$~~ 
Tong Shen${^1}$~~ 
Lianghua Huang${^1}$~~ 
Yu Liu${^1}$~~ 
Yujiu Yang${^2}$
\\
${^1}$Tongyi Lab~~~~~~~~~
${^2}$Tsinghua University
}

\begin{document}
\twocolumn[{%
\renewcommand\twocolumn[1][]{#1}%
\maketitle

\begin{center}
    \centering
    \vspace{-8pt}
    \includegraphics[width=1.0\linewidth]
    {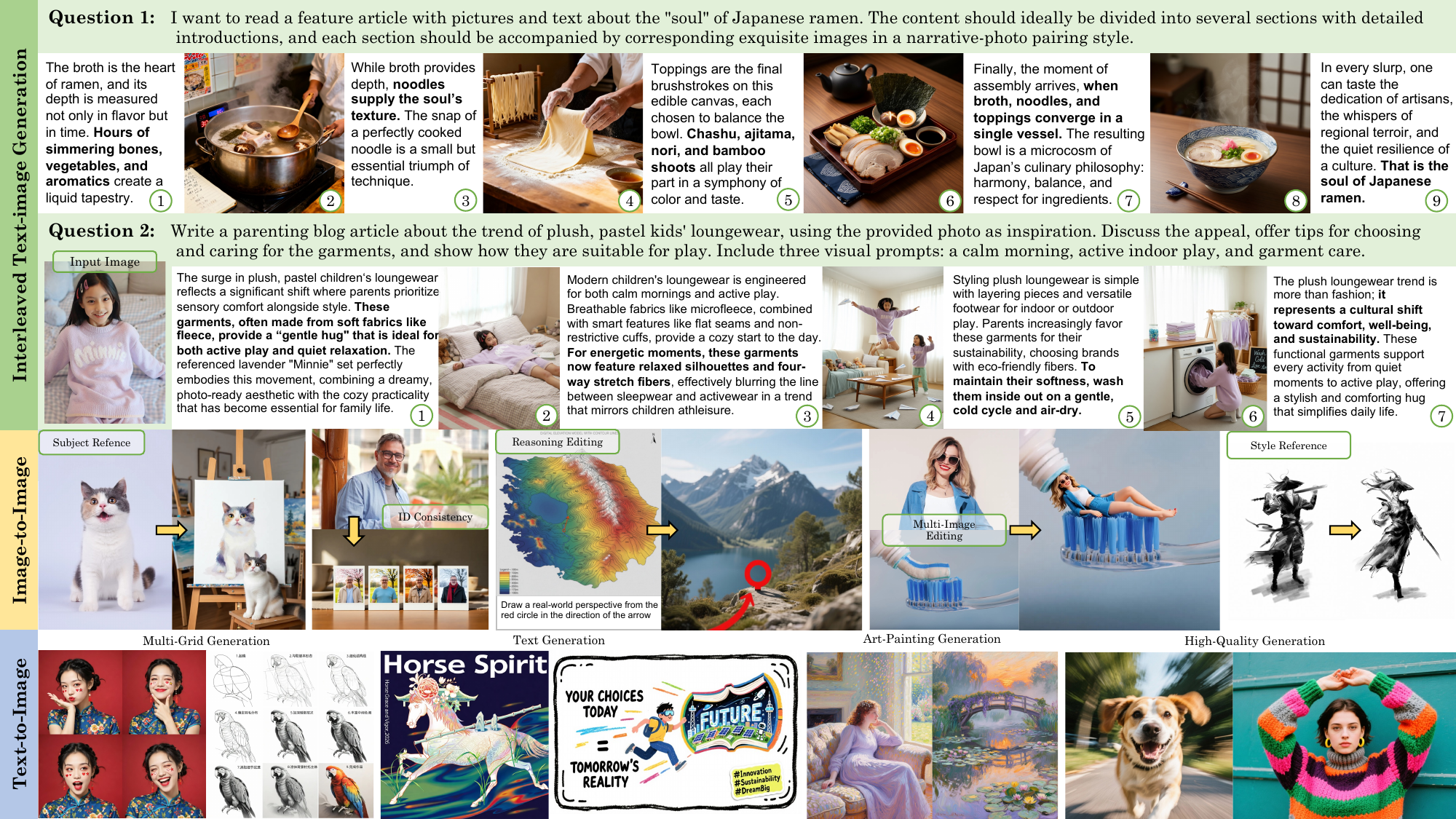}
    \vspace{-19pt}
    \captionsetup{type=figure}
    \caption{Showcase of the versatile abilities of \textit{Wan-Weaver}, including interleaved text-image generation, reference-based image generation/editing, and text-to-image generation.
    }
    \label{fig:teaser}
    \vspace{-2pt}
\end{center}
}]

\let\thefootnote\relax\footnotetext{$^*$Equal contribution.}
\input{sec/0_abstract}

\input{sec/1_introduction}
\input{sec/2_relatedwork}

\input{sec/3_1_method}
\input{sec/4_experiment}

\input{sec/6_conclusion}

\clearpage

\input{sec/X_suppl}
\clearpage
{
    \small
    \bibliographystyle{ieeenat_fullname}
    \bibliography{main}
}

\end{document}

%% file: sec/0_abstract.tex
\begin{abstract}
Recent unified models have made unprecedented progress in both understanding and generation. However, while most of them accept multi-modal inputs, they typically produce only single-modality outputs.  This challenge of producing interleaved content is mainly due to training data scarcity and the difficulty of modeling long-range cross-modal context. To address this issue, we decompose interleaved generation into textual planning and visual consistency modeling, and introduce a framework consisting of a planner and a visualizer. The planner produces dense textual descriptions for visual content, while the visualizer synthesizes images accordingly. Under this guidance, we construct large-scale textual-proxy interleaved data~(where visual content is represented in text) to train the planner, and curate reference-guided image data to train the visualizer. These designs give rise to Wan-Weaver, which exhibits emergent interleaved generation ability with long-range textual coherence and visual consistency. Meanwhile, the integration of diverse understanding and generation data into planner training enables Wan-Weaver to achieve robust task reasoning and generation proficiency. To assess the model’s capability in interleaved generation, we further construct a benchmark that spans a wide range of use cases across multiple dimensions. Extensive experiments demonstrate that, even without access to any real interleaved data, Wan-Weaver achieves superior performance over existing methods.
\vspace{-1pt}
\end{abstract}

%% file: sec/1_introduction.tex
\vspace{-20pt}
\section{Introduction}
\label{sec:introduction}
Generative models are driving progress toward artificial general intelligence. Recent advances in large language models (LLMs)~\cite{touvron2023llama,liu2024deepseek}, vision language models (VLMs)~\cite{bai2025qwen2,Gemini2.5}, unified multi-modal models (UMMs)~\cite{deng2025emerging,chen2025blip3}, and image generation models~\cite{seedream2025seedream,cao2025hunyuanimage,Imagen4} have enabled processing of multi-modal inputs, but usually with single-modal outputs. However, achieving human-like interaction requires the ability to generate multi-turn interleaved multi-modal outputs, which are crucial for applications in reasoning~\cite{mu2023embodiedgpt,byun2024ares}, education~\cite{claman2024artificial,latif2023artificial}, and design~\cite{stella2023can,ko2023large}.

However, generating natural and reliable multi-modal content remains highly challenging, as the outputs across modalities must remain consistent.
Early approaches~\cite{chern2024anole,wu2024next,zheng2023minigpt,kou2024orthus} fine-tune LLMs on image–caption datasets to acquire basic image generation capabilities; however, such training enables only elementary visual synthesis and fails to capture contextual dependencies.
Recent unified multi-modal models (UMMs)~\cite{tian2024mm,team2024chameleon,ge202seedllama} adopt an autoregressive next-token prediction paradigm or its combination with diffusion models, such as BAGEL~\cite{deng2025emerging} and Mogao~\cite{liao2025mogao}, and are pre-trained on interleaved text–image data.
However, the scarcity of large-scale, high-quality interleaved data leads to sparse supervision, making joint optimization unstable and hindering the model’s ability to learn long-range contextual dependencies and modality transitions.

Given the difficulty of data curation, can we achieve interleaved generation without interleaved training data? 
We argue that the goal of \textbf{``interleaved coherence'' could be decomposed into textual, visual, and cross-modal dimensions.}
The textual coherence has already been satisfied by modern VLMs~\cite{bai2025qwen2}. 
The visual coherence among images resembles that
found in reference-based image generation/editing, for which abundant data exist or \textit{can be synthesized}. Likewise, basic text–image alignment in cross-modal coherence can be effectively learned from large text-to-image corpora.
However, \textbf{cross-modal coherence also requires planning capabilities}, \eg, determining where an image should appear in the sequence and what it should portray to fit the long-range context and narrative flow.

In this way, the problem reduces to training a planner and learning each coherence dimension separately.
We therefore propose \textit{Wan-Weaver}, a MoT-architecture~\cite{liang2024mixture,deng2025emerging} unified multi-modal model consisting of a planning expert and a visualization expert.
This design enables decoupled training:
The planner is initialized from a pre-trained VLM, and fine-tuned on large-scale textual-proxy planning and understanding data, where each image is located and represented by a \textit{dense prompt}.
The visualizer is then trained on reference-guided generation data, with the planner kept frozen to provide contextual features. 
During sequential inference,
The planner processes the input together with the previously generated text–image context and produces visualization guidance and plain text, while the visualizer generates corresponding images conditioned on the guidance and visual references. 
Although the dense visual guidance produced by the planner provides useful high-level semantics for the current visualization step, the textual form of the dense prompt inevitably loses subtle contextual cues. Inspired by the notion of context windows~\cite{bai2025qwen2}, we introduce a \textit{dense prompt context window} and further fine-tune the visualizer to improve the consistency of the generated content.

Moreover, our multi-task–trained planner exhibits emergent task reasoning across understanding and generation tasks, advancing toward more intelligent multi-modal generation. To support comprehensive evaluation, we further introduce WeaverBench, a benchmark specifically designed for interleaved generation and covering a broad range of everyday use cases.
Extensive experiments demonstrate that our proposed method not only surpasses leading open-source counterparts but also achieves performance on par with the commercial model Nano Banana. Fig.~\ref{fig:teaser} features our work.
Our contributions are summarized as follows:
\begin{itemize}
    \item We decompose the interleaved multi-modal generation problem and design a unified architecture comprising dedicated planning and visualization experts.
    \item We curate large-scale proxy cross-modal data to address the lack of interleaved supervision and develop a decoupled training strategy that substantially improves interleaved generation over contemporary baselines.
    \item We propose a benchmark for evaluating open-ended interleaved image-text generation, covering a wide range of daily use cases.
\end{itemize}

%% file: sec/2_relatedwork.tex
\section{Related Work}
\label{sec:relatedwork}
\input{figures/overview_interleave}
\paragraph{Unified Multi-modal Models}
~\cite{chen2025janus,ma2025janusflow,wang2024emu3,chen2025blip3,yang2025mmada} aim to build a single architecture capable of both understanding and generation.
They take images and text as input and produce either text or image as output.
While current VLMs generally extend GPT-style LLMs~\cite{brown2020language,radford2018improving,radford2019language} and adopt next-token prediction for multi-modal understanding~\cite{bai2025qwen2,li2024llava,alayrac2022flamingo,zhu2025internvl3,guo2025m2}, while state-of-the-art visual generators rely on diffusion modeling~\cite{esser2024scaling,labs2025flux}. 
This trend has spurred investigations into various paradigms for unifying multi-modal models, which can be roughly classified into three categories: autoregressive (AR) models, fused AR with diffusion models, and diffusion models.
AR models~\cite{sun2023emu,wang2024emu3,sun2024generative,team2024chameleon,wu2024vila,liu2024world,kou2024orthus,yu2023scaling,qu2025tokenflow,fan2025unifluid} follow the paradigm of next-token prediction in language models~\cite{touvron2023llama} by encoding images in continuous embeddings or discrete tokens.
However, AR approaches still lag behind diffusion-based ones in visual fidelity~\cite{chang2022maskgit,li2024autoregressive,tian2024visual,liu2024lumina}.
Fused AR with diffusion models integrates diffusion processes into language backbones, either using additional image decoder~\cite{chen2025blip3,wang2025ovis,ye2024x,tong2025metamorph,pan2025transfer,wu2025openuni,wu2025omnigen2,lin2025uniworld,li2025uniworld,huang2025illume+,wang2025illume,zhang2025nexus,geng2025x,wei2025skywork} or employing a shared Transformer architecture~\cite{xie2024show,xie2025show,zhoutransfusion,shi2024lmfusion,zhao2024monoformer}. Although this hybrid design improves visual quality, it often degrades multi-modal reasoning, as shared parameters must simultaneously optimize for both generation and understanding~\cite{liang2024mixture,lin2024moma}. Recent methods~\cite{deng2025emerging,liao2025mogao} introduce MoT~\cite{liang2024mixture} architecture to successfully achieve better performance with separated Transformer parameters.
Diffusion-based approaches~\cite{yang2025mmada,wang2025fudoki} attempt to completely abandon autoregressive architectures, pursuing unified vision-language modeling from the perspective of discrete diffusion or flow matching, achieving considerable results.
Our unified model adopts the architecture of MoT and employs a decoupled training strategy for interleaved generation.

\vspace{0.5em}
\noindent\textbf{Interleaved Multi-modal Generation}
is a primary use case of unified models~\cite{chen2025janus,ma2025janusflow,ye2024x,koh2023gill,ge202seedllama}. 
Pioneering AR frameworks~\cite{sun2024generative,yu2023scaling} demonstrated that models trained on interleaved sequences could generate both text and images, enabling diverse tasks such as text-to-image generation, visual understanding, and image editing. To enhance the visual realism, subsequent approaches~\cite{zhoutransfusion} substituted image token-wise prediction with iterative denoising for superior image fidelity.
However, these approaches are predominantly designed for single-turn generation, producing an isolated text or image output. 
Recent works~\cite{liao2025mogao,deng2025emerging,kou2024orthus} have attempted to pre-train models on collected interleaved data to enhance their multi-turn multi-modal generation capabilities. However, due to the difficulty in acquiring large-scale and reliable interleaved data, it is challenging for these models to learn complex long-range contextual dependencies. Consequently, this paradigm has yielded suboptimal results. In contrast, we explore a strategy of decoupled training within a unified model, leveraging large-scale synthetic data to achieve superior performance.

%% file: figures/overview_interleave.tex
\begin{figure*}[t]
    \centering
    \includegraphics[width=1\linewidth]{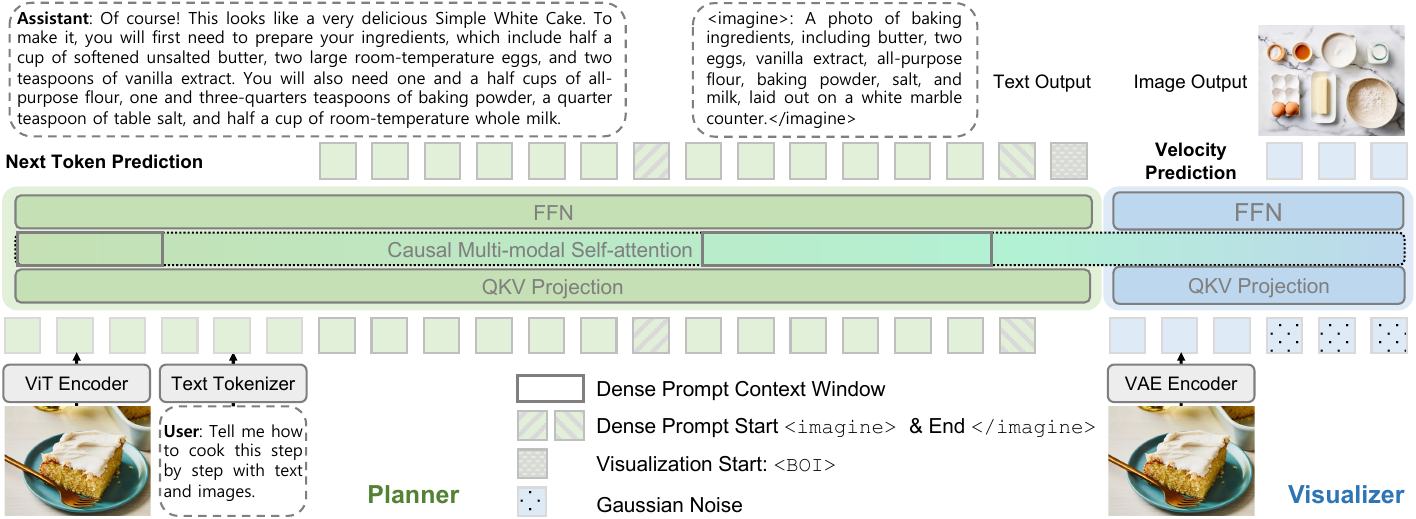} 
    \caption{Overview of the inference process of \textit{Wan-Weaver}. Given a prompt, the planner expert autoregressively generates plain text and dense prompts as visualization cues. Through causal multi-modal self-attention, the visualizer interacts with the planner, enabling it to synthesize images conditioned on the dense prompt context and visual references. The resulting text–image outputs are appended to the history and fed back into the planner, enabling an iterative interleaved generation process that maintains long-range contextual coherence.}
    \label{fig:overview_interleave}
    \vspace{-1em}
\end{figure*}

%% file: sec/3_1_method.tex
\section{Method}
\label{sec:method}

\subsection{Problem Definition}
Interleaved multi-modal generation aims to produce a sequence of text and images from an input prompt. At the modal level, the distribution over this multimodal sequence can be written in a causal form:
\begin{equation*}
\log P_{\theta}(\mathbf{x}) =
\sum_{t=0}^{T}
\log P_{\theta}(\mathbf{x}_{t+1} \mid \mathbf{x}_0, \ldots, \mathbf{x}_t),
\end{equation*}
where $\theta$ denotes the parameters of the model and $\mathbf{x}_{t}$ represents either a text or visual modality rather than an individual token. This formulation is general and encompasses a broad spectrum of existing tasks, including but not limited to text question answering, image captioning, visual question answering, text-to-image generation, and image editing, among others. While our model is also capable of handling these tasks (see Sec.\ref{subsec:single_modality_generation}), they are not the primary focus of this work. Instead, we focus on interleaved text–image generation, with particular emphasis on the more common interleaved pattern, where textual and visual elements are produced alternately in a contextually coherent and semantically aligned manner.

\subsection{Overview of Wan-Weaver}
The overall architecture of Wan-Weaver is illustrated in Fig.~\ref{fig:overview_interleave}. Motivated by our decomposition of interleaved generation into planning and visual coherence modeling, we adopt a unified mixture-of-transformers (MoT)~\cite{liang2024mixture,deng2025emerging} framework comprising a planner and a visualizer expert, which work jointly to produce long-range coherent interleaved text–image content in a coordinated manner.
The planner, instantiated as a VLM, handles modal reasoning and planning, deciding both the modality of the next output and the content of the image when one is required. Given a multi-modal user instruction, text is tokenized and images are encoded with a ViT encoder~\cite{bai2025qwen2}, and the resulting tokens are processed using generalized causal multi-modal self-attention to produce plain text responses. When deciding to transition to the image modality, the planner generates a dense prompt that provides rich visual guidance by aggregating the complex long-range multi-modal context.

The dense prompt triggers the visualizer, implemented as a Diffusion Transformer, which synthesizes the corresponding image. Inspired by context windows in language models~\cite{liu2024deepseek}, we introduce a Dense Prompt Context Window (DPCW) that extracts precise contextual features around the dense-prompt position. The visualizer then performs causal multi-modal self-attention over this window. This targeted conditioning leverages long-range context without introducing excessive noise, leading to better contextual grounding and cross-modal coherence, while VAE-encoded reference tokens help preserve fine details.
Once generated, each text or image is appended to the history for conditioning, enabling Wan-Weaver to autoregressively produce interleaved sequences from any mixture of modalities.

\input{figures/decoupled_training}
\subsection{Decoupled Training Strategy}
\label{subsec:decoupled_train}
Interleaved generation is inherently challenging, as it requires complex cross-modal reasoning over long contexts and demands outputs that are logically consistent, semantically coherent, and aligned in fine-grained details among images.
A straightforward approach is to pre-train a unified model on large-scale interleaved data; however, such high-quality data are scarce,  making the model fail to learn reliable long-range contextual dependencies and modality transitions, and produce misaligned or logically inconsistent cross-modal outputs.
To address this, we decompose interleaved generation into planning and visual coherence and implement a planner–visualizer architecture that supports decoupled training: the planner learns contextual planning from large-scale synthetic textual-proxy and understanding data, while the visualizer learns visual coherence from abundant reference-guided generation data. 
To leverage strong open-source VLMs, we initialize the planner with the pre-trained QWen2.5-VL~\cite{bai2025qwen2}. As no publicly available diffusion transformer matches our visualizer’s architecture, the visualizer is trained from scratch.

\noindent\textbf{Visualizer Training.}
The visualizer is designed to synthesize images aligned with the planner’s visual guidance and to preserve strong reference-driven consistency across images through causal attention.
Achieving this requires substantial generative data. To satisfy the coherence demands of sequential interleaved generation, we decompose the problem into distinct forms of guidance coherence spanning text, single-image, and multi-image contexts.

For text-guidance coherence, the visualizer must align with the semantic space of the planner.
We therefore collect large-scale text–image pairs, including both simple and reasoning-heavy descriptions/instructions, to ensure broad coverage. 
After the first image is generated, the model must also reference previously generated text and images to maintain contextual consistency. 
To this end, we prepare extensive single-image reference data. Similarly, multi-image reference data are incorporated to further enhance long-range consistency across sequences of images. 
Using these coherence-oriented datasets, we freeze the planner and independently train the visualizer with flow-matching loss~\cite{lipmanflow} to achieve multi-modal consistency under diverse conditions, as shown in Fig.~\ref{fig:decoupled_training} (a).

\noindent\textbf{Planner Tuning.}
The planner expert is responsible for digesting complex multi-modal inputs and producing text responses. Crucially, it must infer when to produce text vs. an image and what specific visual content should be generated given the surrounding context—capabilities that our initialized planner only partially supports. Fine-tuning is therefore required, but this is challenging due to the scarcity of large-scale, high-quality interleaved data.

To address this issue, we synthesize large-scale high-fidelity \textit{textual-proxy interleaved data} using top-tier LLMs and VLMs. Each image placeholder is tagged with \texttt{<BOI>} and accompanied by a detailed caption describing the intended visualization.
These fine-grained annotations, called \textit{dense prompts} and enclosed in \texttt{<imagine>}\texttt{</imagine>}, provide richer image-specific guidance than the `sparse' surrounding context and align with the visualizer's textual conditioning during training.

Training with such data offers several advantages:
(1) it leverages the inherent ability of large VLMs to integrate long-range multi-modal information, enabling precise image-generation guidance from purely textual dense prompts and effectively benefiting generation with understanding;
(2) it exploits the semantic equivalence of images and text in the language modeling space, allowing LLMs to learn when to transition modalities without relying on real interleaved data, thus alleviating data scarcity;
and (3) it leads to more stable optimization, as joint-training with denoising generation objectives typically introduces gradient interference~\cite{pan2025transfer}.
Moreover, this textual-proxy mechanism can generalize to other generation tasks, forming textual-proxy generation data. To further equip the planner with automatic task reasoning (\eg, understanding, text-to-image generation, image editing, interleaved generation), we jointly train it on these proxy datasets together with conventional understanding data, as shown in Fig.~\ref{fig:decoupled_training} (b). 

Once trained, our unified model differs from prior UMMs~\cite{deng2025emerging,geng2025x}, which depend on explicit task-specific system prompts or rigid if-else logic to predefine the output modality by users. In contrast, our approach allows the model to implicitly infer task intent from user prompt.

\noindent\textbf{Dense Prompt Context Window Tuning.}
While the generated dense prompt provides a rich textual description of the intended image, it remains a purely text representation and inevitably suffers from information loss, failing to capture subtle contextual nuances. Thus, we introduce the Dense Prompt Context Window (DPCW), a mechanism designed to enhance contextual grounding for image generation based on the generated dense prompt.
Specifically, during the visualization process, a self-attention window is defined on top of the original causal self-attention centered around the position where the dense prompt is produced. Only the contextual information within this window participates in self-attention interactions with the visualizer. Moreover, since the ViT features of the input image are crucial for maintaining semantic faithfulness during generation~\cite{deng2025emerging}, this visual context is also incorporated.

In this way, DPCW not only encapsulates the semantic richness of the dense prompt but also aggregates the \textit{preceding contextual information} accumulated through layer-wise causal self-attention. This results in a more comprehensive context that is continuously propagated across the sequential planning–visualization process, thereby improving the coherence and consistency of the generated content. As illustrated in Fig.~\ref{fig:decoupled_training} (c), we perform an additional DPCW tuning stage to adapt the model to this context-window-aware conditioning. In this stage, only the visualizer is fine-tuned.

\subsection{Data Curation}
\label{subsec:data}

To support the decoupled training strategy, we curate and synthesize large-scale datasets tailored to the objectives of each training stage.

\noindent\textbf{Visualizer tuning data.}
We curated a diverse corpus combining public text-to-image and image editing datasets. The text instructions were augmented through rewriting to simulate various user input scenarios. Beyond standard text–image pairs and public image-to-image datasets, we further construct large-scale image-reference-guided generation data from two sources for visual coherence learning.
First, we extract high-quality key frames from videos and use a VLM to generate detailed textual descriptions as well as instructions describing the changes across selected frame combinations. Because video frames typically exhibit smooth temporal continuity, most variations are localized rather than drastic. As a result, this source mainly provides data for learning fine-grained local edits and high-coherence reference-based generation.
Second, to obtain more general reference-based generation capability that can match that with real interleaved data, we cluster homologous images in our database using SigLIP~\cite{zhai2023sigmoid} and apply a VLM to create structured descriptions spanning both single-image and multi-image reference scenarios. Unlike video frames, these clusters exhibit far richer diversity, including reference-based learning on style, material, pose, expression, and clothing, which provides highly versatile reference supervision.
Together, these datasets enhance the visualizer’s ability to produce consistent and semantically coherent outputs under diverse reference conditions.

\input{tables/main_compare_opening}
\noindent\textbf{Planner tuning data.}
To preserve the language modeling capacity of the underlying VLM, we collect high-quality text-only and image–text understanding data. To endow the planner with interleaved generation planning capability, we constructed \textit{textual-proxy data}, where visual content was replaced by detailed captions annotated with \texttt{<imagine>}. Specifically, three types of data were synthesized: (1) large-scale user query (text only)–interleaved article pairs generated by prompting the top-tier LLMs with category or tag keywords; (2) user query (with images)–interleaved article pairs built around arbitrary images from the database by VLMs; and (3) multi-image data where each image was first captioned by a VLM, then organized into coherent interleaved narratives, and finally refined for logical and stylistic consistency. This refinement step is essential, as independently generated VLM captions may introduce inconsistencies due to inherent sampling variability. Note that the dense prompts in the textual-proxy data exhibit substantial diversity: they range from short phrases to long, detailed descriptions, and may also specify changes relative to particular reference images.
The same proxy strategy was also applied to non-interleaved generation tasks such as text-to-image and reference-based image generation, improving the reasoning and user intent comprehension ability.

\input{figures/benchmark_stat}

%% file: figures/decoupled_training.tex
\begin{figure}[t]
    \centering
    \includegraphics[width=1\linewidth]{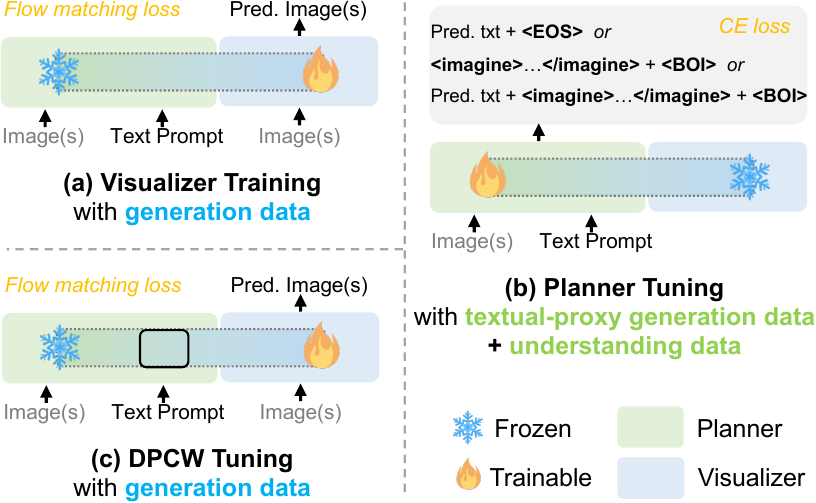} 
    \caption{Illustration of our decoupled training strategy.}
    \label{fig:decoupled_training}
    \vspace{-1em}
\end{figure}

%% file: tables/main_compare_opening.tex
\begin{table*}[htbp]
    \caption{Quantitative comparison with existing state-of-the-art methods on interleaved generation benchmark OpenING~\cite{zhou2025opening}.}
    \vspace{-0.5em}
    \label{tab:leaderboard}

    \resizebox{\linewidth}{!}{
    \setlength{\tabcolsep}{2.1pt}
    \centering
    \begin{tabular}{lccccccc|c}
        \toprule
        Method  & Completeness & Quality & Richness & Correctness & Human Alignment & IT Coherency & 
        Multi-step Consistency & Overall\\
        \midrule
        
        NExT-GPT  & 3.89 & 4.25 & 3.35 & 3.61 & 5.35 & 3.32 & 3.85 & 3.95\\ 
        MiniGPT-5  & 3.91 & 4.50 & 3.61 & 3.63 & 5.51 & 3.56 & 4.10 & 4.12\\ 
        Orthus & 4.43& 4.30& 3.71& 4.15& 4.80& 3.51 & 4.20 & 4.16\\
        Show-o  & 4.37 & 4.79 & 3.83 & 3.76 & 5.78 & 4.04 & 4.33 & 4.41\\
        VILA-U  & 5.60& 5.14& 4.68& 4.78& 5.69& 4.74& 4.79& 5.06\\ 
        SEED-LLaMA  & 5.59 & 5.50 & 4.61 & 4.59 & 6.50 & 4.43 & 5.13 & 5.19\\ 
        Anole  & 6.27 & 6.02 & 5.28 & 5.06 & 6.91 & 4.90 & 5.81 & 5.75\\
        Emu3  & 5.90& 5.96& 5.52& 5.43& 6.47& 5.66& 5.37& 5.76\\ 
        SEED-X  & 5.65 & 6.07 & 4.92 & 5.77 & 7.03 & 5.72 & 5.72 & 5.84\\ 
        Gemini+Flux  & 7.58 & 7.26 & 6.48 & 7.03 & 7.98 & 6.98 & 7.33 & 7.23\\ 
        GPT-4o+DALL-E3  & 8.66 & 8.01 & 7.42 & 7.98 & \seco{8.77} & 8.15 & 8.38 & 8.20\\ 
        Nano Banana  & \seco{9.34}& \best{8.58}& \seco{8.00}& \best{9.17}& \best{8.88}& \best{9.27}& \best{8.70}& \best{8.85}\\ 
        \rowcolor{Gray}
        Wan-Weaver (Ours)  &\best{9.41}& \seco{8.32}& \best{8.03}& \seco{8.90}& 8.69& \seco{8.78}& \seco{8.56}& \seco{8.67}\\ 
        \bottomrule
    \end{tabular}
    
}
\vspace{-1.5em}
\end{table*}

%% file: figures/benchmark_stat.tex
\begin{figure}[t]
    \centering
    \includegraphics[width=1\linewidth]{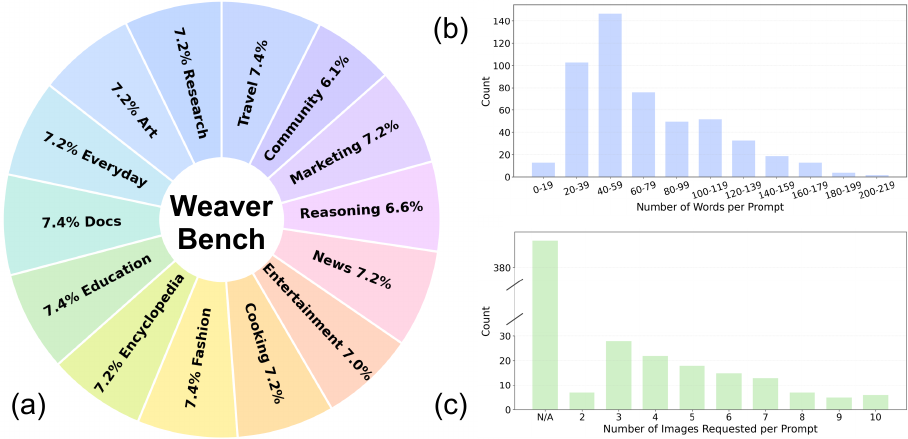} 
    \caption{Statistics of WeaverBench. (a) Topic distribution across 14 everyday categories. (b) Prompt length distribution. (c) Distribution of the number of images requested per prompt.}
    \vspace{-1.5em}
    \label{fig:benchmark_stat}
\end{figure}

%% file: sec/4_experiment.tex
\section{Experiment}
\label{sec:experiment}
\subsection{Implementation Details}
We initialize the planner with an in-house Qwen2.5-VL-32B-Think~\cite{bai2025qwen2} model and implement a twin-structured, DiT-based visualizer, trained from scratch. A frozen visual encoder—VAE from Wan2.2~\cite{wan2025wan} plus the Qwen2.5-VL ViT—is used throughout all experiments. The visualizer is trained for 9.6T tokens with AdamW optimizer, using a learning rate that decays from $5\times10^{-5}$ to $2.5\times10^{-5}$; images keep native aspect ratios with resolutions spanning $\sim$196$^2$ to 1440$^2$, and the share of high-resolution data is progressively increased. Training alternates over three stages: text-to-image, text-image-to-image, and text-multi-image-to-image. The planner is tuned over 35.72G tokens with AdamW at $7\times10^{-6}$, using a 5:1 generation-to-understanding sampling ratio. DPCW is realized via an attention-masking strategy, and 3D RoPE~\cite{wan2025wan} is employed.

\subsection{Benchmark: WeaverBench}
Prior benchmarks~\cite{an2024openleaf,liu2024holistic} target only a narrow set of topics. While recent efforts aim for broader coverage, some of their user prompts nevertheless involve non-interleaved tasks (\eg, pure understanding or single-image editing)~\cite{zhou2025opening} or remain overly templated~\cite{chen2024interleaved}. We therefore introduce a compact, interleaved-generation–focused benchmark, characterized by flexible user queries and diverse generation scenarios grounded in everyday use, aiming to foster more rigorous and practically meaningful assessment.

With the assistance of multiple AI agents, we collaboratively brainstormed and identified a comprehensive set of everyday scenarios that demand interleaved image–text generation. These insights were consolidated into 14 primary categories. To construct corresponding test prompts, we manually collected and crafted user queries from diverse sources (\eg, social media, search engines, and public knowledge bases) to ensure broad topical coverage. The resulting prompts exhibit varying levels of specificity, ranging from short queries with only a few words to complex requests that specify the number, order, and semantic content of each image to be generated. In total, the benchmark comprises 512 test cases, evenly split between image-conditioned and text-only prompts. Fig.~\ref{fig:benchmark_stat} presents the statistics of our WeaverBench.

\input{figures/nano_compare}
We carefully identify the primary aspects of interleaved generation and design corresponding evaluation metrics, including Prompt Adherence, Narrative Coordination, Content Consistency, and Completeness. As demonstrated in~\cite{zhou2025opening}, GPT-4o~\cite{openai2025gpt4o} aligns well with human preferences. Following this, we develop finer-grained scoring rules to obtain more reliable assessments.

\subsection{Interleaved Image-Text Generation}
\input{tables/Weaver_bench}
To evaluate the quality of interleaved generation, we employ the OpenING benchmark~\cite{zhou2025opening}, a dataset comprising over 2,000 samples, along with the seven metrics proposed in the original paper. We evaluate our method against a broad range of existing approaches, which can be categorized into four groups. (1) Integrated pipelines combine independent text and image generation models: GPT-4o+DALL$\cdot$E-3~\cite{openai2025gpt4o,betker2023improving} and Gemini+Flux~\cite{team2023gemini,labs2025flux}. (2) Two-stage generators that adopt a unified model architecture but generate textual and visual content sequentially in two distinct stages: Emu3~\cite{wang2024emu3}, SEED-X~\cite{ge2024seed}, VILA-U~\cite{wu2024vila}, and Show-o~\cite{xie2024show}. (3) End-to-end generators produce interleaved image–text content within a single autoregressive process: 
Orthus~\cite{kou2024orthus}, NExT-GPT~\cite{wu2024next}, MiniGPT-5~\cite{zheng2023minigpt}, SEED-LLaMA~\cite{ge202seedllama}, and Anole~\cite{chern2024anole}. (4) Commercial interleaved generators that directly support text–image interleaving: Gemini-2.5-Image (Nano Banana)~\cite{google2025nano}. Table~\ref{tab:leaderboard} shows that our method clearly surpasses all open-source and integrated-pipeline baselines. While the top-tier commercial model Nano Banana—whose implementation details are not publicly available—retains a small overall lead, our method is highly competitive and outperforms it on several metrics. We further note that Nano Banana's strong image–text coherence and multi-step consistency may arise from repeated or highly similar image outputs, as illustrated in Fig.~\ref{fig:nano_compare}.
In addition, we further assess representative unified models (with general interleaved capabilities~\cite{kou2024orthus,chern2024anole,cui2025emu35,google2025nano}) on our WeaverBench to examine performance across diverse daily use cases. As shown in Table~\ref{tab:weaver_bench_results}, the results reveal a similar conclusion, further demonstrating the superior performance of our method.

\input{tables/single_modality}

\subsection{Single Modality Generation}
\label{subsec:single_modality_generation}

Our model is a unified multi-modal generation framework trained on diverse understanding and generation data, enabling it to support both interleaved multi-modal generation and standard single-modality tasks. For comparison, we include understanding-only models (\ie, InternVL3-38B~\cite{zhu2025internvl3}, Ovis2-34B~\cite{lu2025ovis2}, and Qwen2.5-VL-32B~\cite{bai2025qwen2}), generation-only models (\ie, FLUX.1-dev~\cite{labs2025flux} and Step1X-Edit~\cite{liu2025step1x}), and unified models (\ie, BAGEL~\cite{deng2025emerging} and UniWorld-V1~\cite{lin2025uniworld}). As shown in Table~\ref{tab:single_comparison}, it delivers strong performance across understanding, image generation, and editing benchmarks, markedly outperforming previous unified and specialized generation models. The qualitative results are shown in Fig.~\ref{fig:teaser}.

\subsection{Ablation Studies}
We perform the ablation studies using the 7B variant owing to computational limitations.

\noindent\textbf{Decoupled Training.}
To study the superiority of our decoupled training strategy, we compare it with the naive joint-training strategy on the full datasets. As shown in Fig.~\ref{fig:vision_loss}, where P+V indicates joint-training of the planner and visualizer, and V (T2I+SI2I+MI2I) denotes visualizer tuning, the decoupled training setting steadily reduces the vision loss from around 0.25 to 0.15 and yields a much smoother optimization trajectory. This behavior is expected: when the planner and visualizer are optimized jointly, the non-negligible semantic and distribution gap between textual and visual modalities introduces instability, leading to misaligned text–image generation and degraded image quality. In contrast, isolating visualizer training avoids this cross-task interference, resulting in more stable convergence.

\input{figures/vision_loss}

\input{figures/ablation_planner}

\noindent\textbf{Feature Modeling in Planner.} 
To investigate how planning levels affect understanding capability, we perform planner tuning on various data compositions. The baseline is a VLM-initialized planner without planning ability. As shown in Fig.~\ref{fig:ablation_planner}, we compare three variants: (1) \textbf{+und.\&gen.\&proxy} provides comprehensive planning including dense prompts; (2) \textbf{+und.\&gen.} preserves basic understanding with generative planning; and (3) \textbf{+und.} maintains only understanding data.

Fig.~\ref{fig:ablation_planner} (left) shows that understanding performance remains stable across configurations, indicating that planning does not compromise core understanding competency. We further evaluate modality-specific patterns via structural plan token accuracy. The planner must correctly emit \texttt{<BOI>} tokens: none for understanding, exactly one for T2I/I2I, and at least one for interleaved generation. As shown in Fig.~\ref{fig:ablation_planner} (right), without generation data, the model fails to emit image signals. However, as the generation-oriented data ratio increases (from 1g1u to 5g1u), planning proficiency improves substantially, with a gradual increase in image starting tokens for interleaved tasks. Balancing planning reliability and understanding stability, we adopt the 5g1u ratio as the final composition for planner tuning.

\noindent\textbf{Coherence Modeling in Visualizer.} 
We constructed several training variants of our visualizer to investigate the effectiveness of our coherence modeling. Specifically, we train the visualizer using three settings:
(1) only text–image paired data (T2I data),
(2) T2I combined with single-image-to-image data (T2I+SI2I data), and
(3) multi-image-to-image data added on top of all previous data (T2I+SI2I+MI2I data), which is our strategy.
The corresponding qualitative result is shown in Fig.~\ref{fig:ablation_data_imagine}. We can see that T2I data-only training provides basic text–image alignment but lacks any reference ability, failing to generate the second image. Adding single-image reference data improves appearance preservation across steps, while introducing multi-image reference data further strengthens long-range visual coherence, enabling consistent object identity, style, and detail across multiple generated images.

As shown in Fig.~\ref{fig:ablation_data_imagine} (right), the model without the dense prompt tends to generate repetitive and contextually irrelevant images. In contrast, using the dense prompt results in significantly more diverse outputs. The integration of our DPCW further enhances the model's ability to adhere to user instructions, particularly in long-context scenarios.

\input{figures/ablation_visual}

%% file: figures/nano_compare.tex
\begin{figure*}[t]
    \centering
    \includegraphics[width=\linewidth]{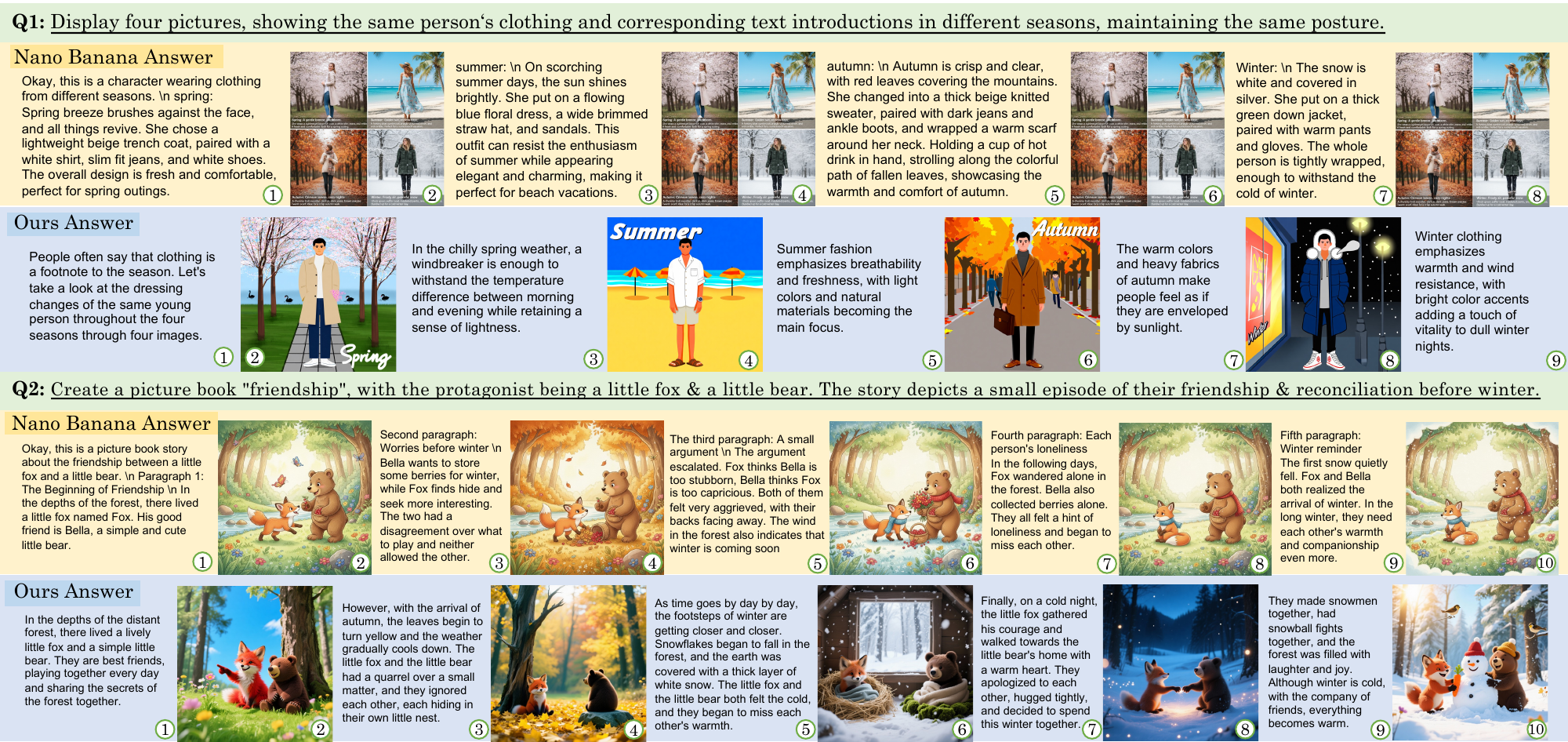}
    \vspace{-1em}
    \caption{Qualitative comparison with the state-of-the-art commercial system Nano Banana on interleaved text–image generation.}
    \vspace{-1em}
    \label{fig:nano_compare}
\end{figure*}

%% file: supp/weaver_bench.tex
\begin{table*}[!t]
    \caption{Quantitative comparison on WeaverBench. PA: Prompt Adherence, NC: Narrative Coordination, CC: Content Consistency, IC: Image Consistency, CP: Completeness, Acc: Accuracy of the number of the generated images.}
    \label{tab:weaver_bench_results_supp}

    \resizebox{\linewidth}{!}{
    \centering
    \begin{NiceTabular}{lcccccccc|c||c}
        \toprule
\multirow{2}{*}{Method} &
\multirow{2}{*}{PA} &
\multirow{2}{*}{NC} &
\multicolumn{3}{c}{CC} &
\multicolumn{2}{c}{IC} &
\multirow{2}{*}{CP} &
\multirow{2}{*}{Overall} &
\multirow{2}{*}{Acc} \\
\cmidrule(lr){4-6} \cmidrule(lr){7-8} 
          & & & Intra-output & Input-to-image & Input-to-text & Entity & Style &  &  & \\
        \midrule
        Orthus & 2.47& 1.88& 1.42 & 1.30 & 2.35 & 1.19& 1.82& 1.91& 1.89 & 3.28\%\\
        Anole & 4.14& 3.76& 3.56& 3.31& 4.44& 3.00& 3.83 &3.64 &3.74 & 6.56\%\\
        Emu3.5 & 7.65& 7.55& 7.42& 7.41& 7.86& 7.22& 7.78& 7.41& 7.53& 48.36\% \\
        Nano Banana & \seco{8.53}& \seco{8.19}& \best{8.31}& \best{8.57}& \seco{8.70} & \best{8.09} & \best{8.67}& 8.29&\seco{8.38} & \seco{66.39}\%\\
        \rowcolor{Gray}
        Wan-Weaver (Ours) &\best{8.71} &\best{8.33} & \seco{8.26} &\seco{8.41} & \best{8.83} & \seco{7.85}& \seco{8.41}&\best{8.46} & \best{8.43}& \best{93.44}\%\\
        \bottomrule
    \end{NiceTabular}
}
\end{table*}

%% file: tables/single_modality.tex
\begin{table}[!t]
\centering
\caption{Comparison across single-modality generation tasks (understanding, image generation, and editing). $^{\dagger}$: Our in-house base model with thinking mode (enabled only for understanding).}
\vspace{-0.5em}
\label{tab:single_comparison}
    \resizebox{\linewidth}{!}{
\setlength{\tabcolsep}{2.1pt}
\begin{tabular}{l cc cc cc}
\toprule
\multirow{2}{*}{Model} &
\multicolumn{2}{c}{\textbf{Understanding}} &
\multicolumn{2}{c}{\textbf{Image Generation}} &
\multicolumn{2}{c}{\textbf{Image Editing}} \\
\cmidrule(lr){2-3} \cmidrule(lr){4-5} \cmidrule(lr){6-7}
& MMMU & MathVista & GenEval & DPG & ImgEdit & GEdit--EN \\
\midrule
InternVL3-38B          & 69.7 & 76.3 & --   & --    & --   & --   \\
Ovis2-34B              & 66.7 & 76.1 & --   & --    & --   & --   \\
Qwen2.5-VL-32B$^{\dagger}$         & \best{75.1} & \best{84.7} & --   & --    & --   & --   \\
\midrule
FLUX.1-dev             & --   & --   & 0.66 & 84.0  & --   & --   \\
Step1X-Edit            & --   & --   & --   & --    & 3.06 & \seco{6.70} \\
\midrule
\textbf{Unified Models} & & & & & & \\
Bagel                  & 55.3 & 73.1 & \seco{0.88} & \seco{85.07} & 3.20 & 6.52 \\
UniWorld-V1            & 58.6 & --   & 0.84 & 81.38 & \seco{3.26} & 4.85 \\
\rowcolor{Gray}
Wan-Weaver (Ours)                    & \seco{74.9} & \seco{84.3} & \best{0.89} & \best{87.21} & \best{4.31} & \best{7.39} \\
\bottomrule
\end{tabular}
}
\vspace{-1em}
\end{table}

%% file: figures/vision_loss.tex
\begin{figure}[t]
    \centering
    \includegraphics[width=0.8\linewidth]{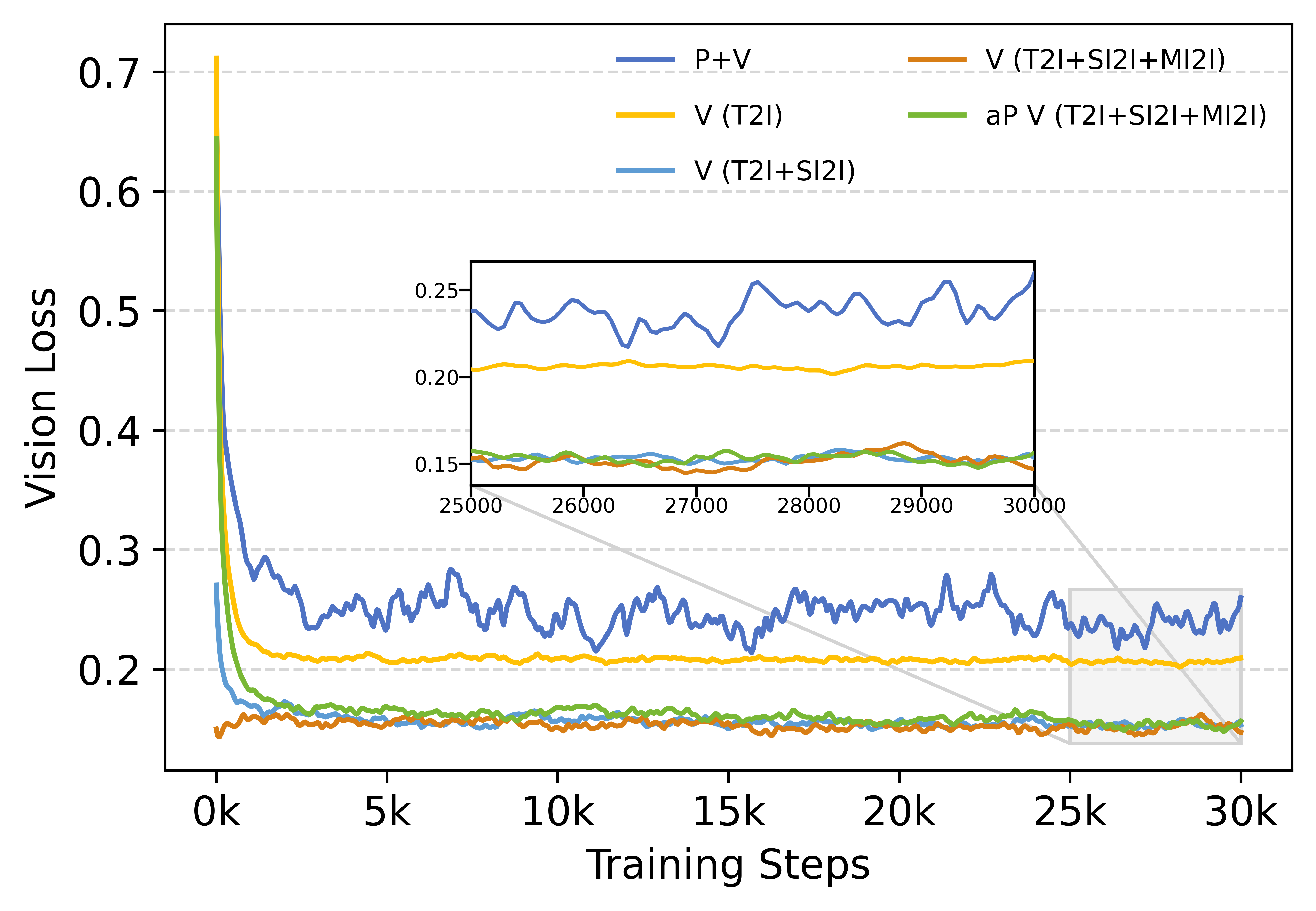} 
    \vspace{-0.5em}
    \caption{Loss curves of different training strategies.}
    \label{fig:vision_loss}
    \vspace{-0.5em}
\end{figure}

%% file: figures/ablation_planner.tex
\begin{figure}[t]
    \centering
    \includegraphics[width=1\linewidth]{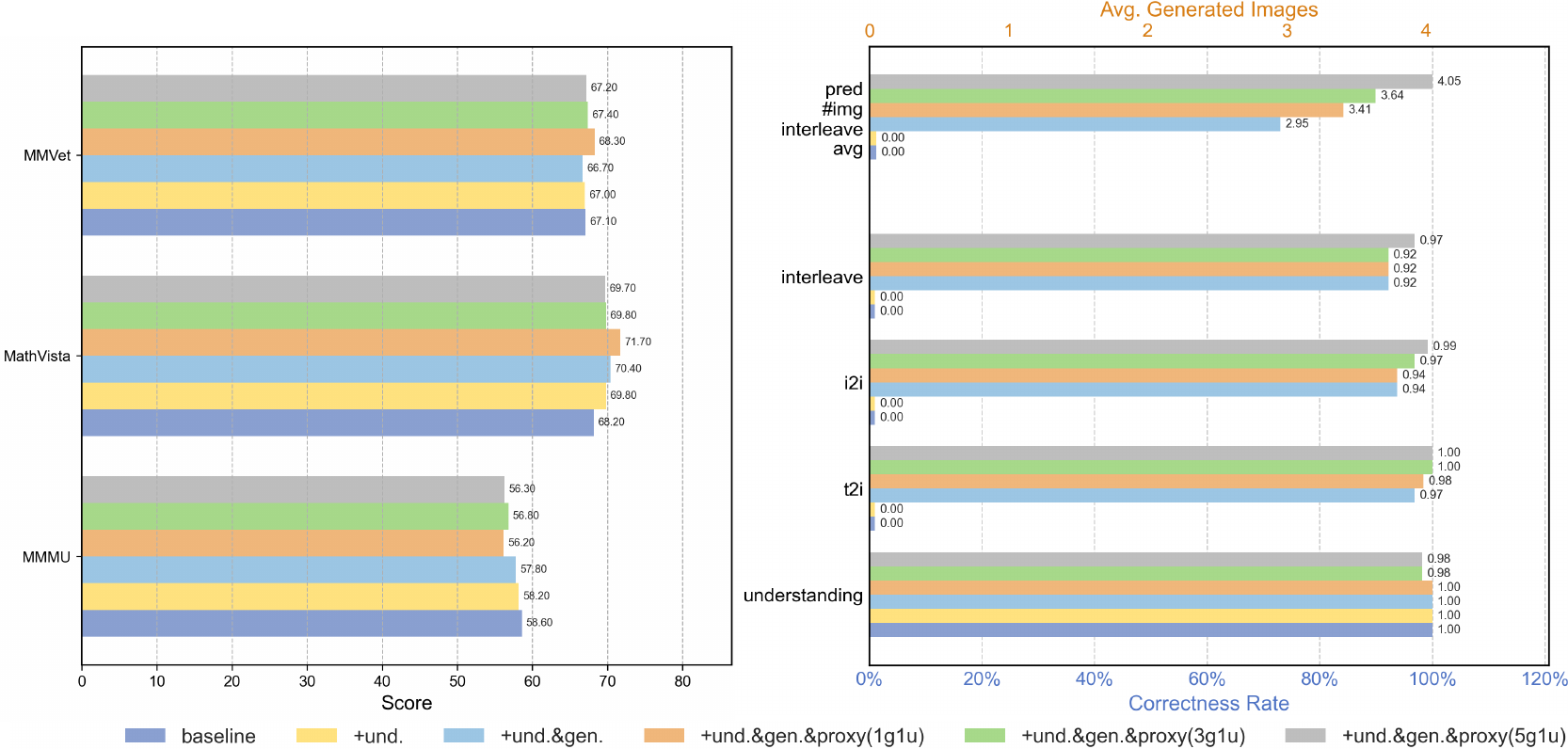} 
    \vspace{-1.5em}
    \caption{Impact of training on different types of data on our planner. (left) Performance on understanding metrics. (right) Token prediction statistics per task.}
    \vspace{-1em}
    \label{fig:ablation_planner}
\end{figure}

%% file: figures/ablation_visual.tex
\begin{figure}[t]
    \centering
    \includegraphics[width=\linewidth]{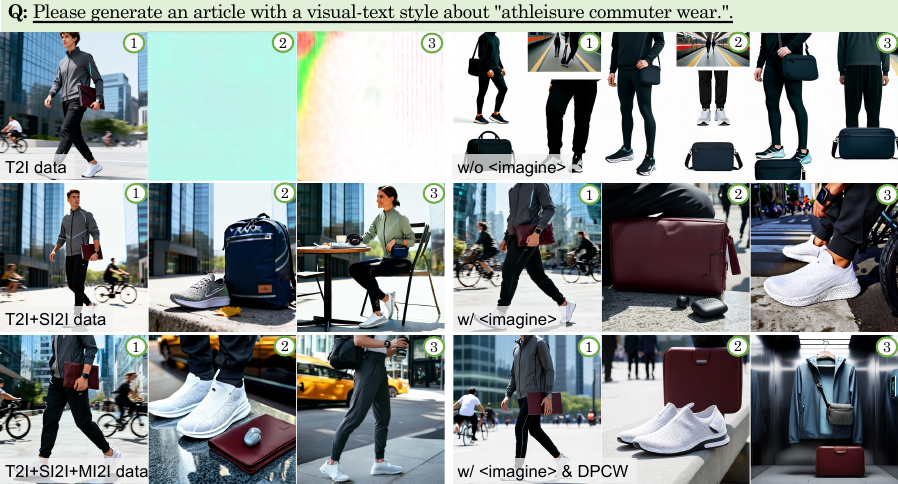} 
    \vspace{-1.3em}
    \caption{Visual comparison of the results generated by different variants of our method.}
    \label{fig:ablation_data_imagine}
    \vspace{-1.2em}
\end{figure}

%% file: sec/6_conclusion.tex
\section{Conclusion}
\label{sec:conclusion}
We presented Wan-Weaver, a unified multi-modal model with planner-visualizer architecture for interleaved text–image generation under limited interleaved supervision. By decomposing interleaved coherence into planning and visual coherence, synthesizing large-scale textual-proxy interleave data, and adopting a decoupled learning strategy, it effectively learns long-range contextual dependencies and generates consistent multi-modal content. Extensive experiments demonstrate our model substantially outperforms existing open-source approaches and delivers performance competitive with top-tier commercial demos.

%% file: sec/X_suppl.tex
\clearpage
\setcounter{section}{0}
\maketitlesupplementary
\appendix

\tableofcontents
\vspace{4mm}
More showcases and visual comparisons are available at \url{https://doubiiu.github.io/projects/WanWeaver}.

\section{Societal Impact}
\label{sec:impact}
\textit{Wan-Weaver}, an innovative interleaved text–image generation technology, has the potential to substantially advance human–AI interaction by enabling richer, more natural, and more expressive communication. A unified model capable of producing coherent mixed-modality content can support a wide range of beneficial applications, including educational content creation, visual tutoring systems, accessibility tools that jointly describe and illustrate concepts, and collaborative design workflows where users iteratively steer both text and images. These capabilities may also reduce the entry barrier for creative industries by assisting individuals without professional illustration or design experience in producing high-quality visual materials.

However, as with other large-scale generative models, interleaved generation systems carry inherent societal risks. Because they are trained on large, heterogeneous datasets, such models may inherit and amplify biases present in the underlying corpora, potentially leading to stereotypical or culturally insensitive generations across both textual and visual modalities. The ability to generate tightly interwoven text–image content also increases the risk of producing persuasive misinformation or synthetic narratives that appear more credible due to coherent visual–textual alignment. Moreover, without appropriate safeguards, the model may generate harmful, inappropriate, or misleading visual content when prompted adversarially. 

Our project is research-oriented and focuses on the scientific exploration of unified interleaved generation models. Nonetheless, we emphasize the importance of responsible deployment: future practical systems should incorporate robust content filters, watermarking, auditing mechanisms, and human-in-the-loop controls to mitigate potential misuse. Continued study of bias in multi-modal generation remains crucial as such technologies evolve.

\begin{table*}[!t]
    \caption{Distribution of data samples across the 15 categories in the WeaverBench benchmark. (\textbf{The number before ‘+’} represents the amount for text-only input, and \textbf{the number after ‘+’} represents the amount for image-text input.)}
    \label{tab:data_sample}
    \vspace{-0.5em}

    \resizebox{\linewidth}{!}{
    \centering
    \begin{tabular}{lccccc}
    \toprule
    Category & Encyclopedic Knowledge & News Media & Travel Guide & Daily Life & Food Cooking   \\
    \#Data Samples & \textbf{35} (16+19) & \textbf{35} (16+19) & \textbf{36} (17+19) & \textbf{35} (16+19) & \textbf{35} (16+19) \\

    \midrule
    Category & Education Tutorial & Art Creation & Fashion Beauty & Media Entertainment & Product Marketing \\
    \#Data Samples  & \textbf{36} (17+19) & \textbf{35} (16+19) & \textbf{36} (17+19) & \textbf{34} (16+18) & \textbf{35} (16+19)\\
    
    \midrule
    Category & Academic Research  & Document Guides & Social Community & Reasoning & Multi-language\\
    \#Data Samples & \textbf{35} (17+18) & \textbf{36} (17+19) & \textbf{29} (15+14) & \textbf{32} (16+16) & \textbf{28} (28+0) \\
    
    \bottomrule
    \end{tabular}
}
\end{table*}

\section{Details of Our Benchmark: WeaverBench}
\label{sec:benchmark}

\subsection{Category Definition}
In this section, we present the definitions of 15 distinct categories within the WeaverBench, an evaluation framework designed for comprehensive performance assessment across various domains. Notably, a special `multi-language' category also exists, which is integrated across 14 content-oriented categories. The precise definitions ensure that the benchmarks are relevant and reflective of real-world tasks. These categories are selected to encompass a wide range of familiar and frequently encountered challenges, ensuring an invaluable tool for evaluating models across diverse contexts.

\textbf{Encyclopedic Knowledge.} 
This category includes prompts related to articles or information found in encyclopedias, encompassing a broad spectrum of fundamental knowledge, concepts, and factual data. It emphasizes the model's ability to accurately retrieve, synthesize, and present encyclopedic content.

\textbf{News Media.}
Focused on the domain of news and media, this category involves tasks such as creating news reports, offering timely commentary, media analysis, and understanding of social dynamics. Models are tested on their capability to generate content that is informative and adheres to the stylistic norms of professional journalism.

\textbf{Travel Guide.}
Involving destination recommendations, travel tips, itinerary planning, and cultural background information, this category assesses the ability to generate informative and engaging content for travel enthusiasts.

\textbf{Daily Life.}
This category encompasses everyday tasks and challenges, including vlogs, household management, wellness advice, financial guidance, and interpersonal relationship tips. It measures a model's proficiency in simulating real-world day-to-day scenarios and providing practical solutions.

\textbf{Food Cooking.}
Tasks under this category involve recipes, cooking techniques, culinary exploration, and restaurant reviews. Content generation assessed here requires models to deliver accurate, creative, and user-friendly culinary insights.

\textbf{Education Tutorial.}
This category involves developing educational materials such as online courses, study guides, lecture notes, and learning strategies. The ability to generate educational content that is pedagogically sound and engaging to learners is key.

\textbf{Art Creation.}
Encompassing prompts related to photography, painting, drawing, handicrafts, art techniques, and exhibition of works, this category assesses creativity. Models are evaluated on their ability to produce content that reflects artistic insight and innovation.

\textbf{Fashion Beauty.}
This category pertains to fashion trends, makeup techniques, skincare advice, and personal style. It evaluates the model's ability to provide up-to-date fashion insights and beauty tips in line with industry trends.

\textbf{Media Entertainment.}
Focused on film, television, variety show reviews, recommendations, and related information, this category assesses a model's comprehension and ability to engage audiences in the realm of entertainment media.

\textbf{Product Marketing.}
Involving product introductions, campaign strategies, sales techniques, industry trends, and advertising case studies, this category is designed to test the model's capability to simulate market-oriented content generation and promotional strategies.

\textbf{Academic Research.}
This category covers academic papers, research developments, scholarly discussions, and related resources. It assesses the model's proficiency in generating well-founded and articulate academic content that meets the rigors of scholarly work.

\textbf{Document Guides.}
Involves the creation of technical documents, user manuals, operating guides, and practical learning materials. The focus here is on clarity and utility in the generation of instructive content.

\textbf{Social Community.}
This category is centered on user interaction, forum discussions, social media dynamics, and community activities. Models are evaluated on their capacity to generate engaging and constructive content that fosters community interactions.

\textbf{Reasoning.}
In this category, prompts require complex, multi-step reasoning to generate answers. Tasks are designed to test a model's cognitive capabilities in logical deduction, problem-solving, and articulation of coherent arguments.

\textbf{Multi-language.}
This category primarily consists of English prompts that instruct the model to generate images containing text written in other languages. This task evaluates the model's proficiency in handling multilingual contexts and integrating textual and visual data.

Through the detailed exploration of these categories, WeaverBench provides a rigorous and holistic framework for evaluating and improving the capabilities of current state-of-the-art models. We list the number of data samples in each category in Table~\ref{tab:data_sample}.

\input{supp/Weaver_bench}
\subsection{Evaluation Metrics}
\label{subsec:supp:evaluation_metrics}
\textbf{Prompt Adherence} measures the extent to which the generated image–text output faithfully follows the user’s instruction and fulfills the specified task. This metric captures not only topical relevance, but also the correctness and usefulness of the solution provided with respect to all explicit and implicit requirements encoded in the prompt. High Prompt Adherence implies that the system addresses every key aspect of the request, avoids digressions into irrelevant content, and produces responses that are factually accurate and operationally helpful for the intended task. Conversely, low adherence is characterized by missing or ignoring parts of the instruction, addressing a different or incomplete task, or producing fundamentally incorrect or misleading content. In practice, this metric evaluates whether the output is directly responsive to the prompt, covers the necessary subcomponents, and provides an answer that a user could plausibly adopt to accomplish the stated goal.

\textbf{Narrative Coordination} assesses how effectively text and images are organized and sequenced to form a coherent, comprehensible multimodal narrative. It focuses on whether images appear at appropriate points in the response to support or clarify the corresponding textual explanation, and whether the overall progression across modalities facilitates step-by-step understanding. High Narrative Coordination is characterized by a deliberate alignment between modalities, in which visual elements are introduced at conceptually critical junctures, reinforce or concretize key textual descriptions, and avoid redundancy or disruption of the reading flow. In contrast, poor coordination may manifest as images that are placed arbitrarily, lack necessary accompanying text, appear out of order relative to the described steps, or introduce confusion rather than clarity. This metric thus captures the structural quality of the multimodal presentation, emphasizing pedagogically sound timing, logical ordering, and smooth integration of textual and visual information.

\textbf{Content Consistency} evaluates the degree of semantic and stylistic alignment between images, text, and the user’s original input. It consists of three complementary aspects: (1) Intra-output Coherence, which examines whether each generated image matches the adjacent textual description in content, mood, and style (e.g., an image labeled as “a watercolor sketch of a rainy street” should visually resemble a watercolor depiction of a rainy urban scene); (2) Input-to-Image Fidelity, which measures how well the generated images respect the explicit and implicit visual constraints specified in the user’s prompt, such as character attributes, scene configurations, or requested artistic styles; and (3) Input-to-Text Fidelity, which assesses whether the generated text adheres to the intended domain, tone, and factual expectations of the task (for example, employing precise, formal exposition for scientific explanations rather than casual or poetic language). High Image–Text Consistency indicates that images and text are mutually supportive, accurately reflect the user’s constraints, and maintain a coherent semantic and stylistic relationship throughout the output.

\textbf{Image Consistency} measures the stability and coherence of visual and semantic elements across multiple steps, scenes, or turns within a multi-image or multi-stage output. This metric comprises two main dimensions: (1) Entity Consistency, which evaluates whether recurring objects, characters, or scenes maintain identifiable and stable properties across all relevant images, including aspects such as color, shape, clothing, and spatial relationships. Substantial unexplained changes in these properties reduce the user’s ability to track entities and undermine narrative coherence. (2) Visual Style Uniformity, which assesses the extent to which the images share a consistent visual style, such as medium (e.g., illustration vs. photograph), color palette, lighting conditions, contrast, and overall aesthetic. High Multi-step Consistency implies that the output resembles a cohesive visual sequence (e.g., frames from the same story or instructional series), allowing users to easily follow longitudinal processes or narratives without being distracted by unintended variations in entity appearance or visual style.

\textbf{Completeness} evaluates whether the multimodal output covers all essential components, steps, or stages implied or explicitly required by the user’s request, and whether the provided explanations and visual supports are sufficiently detailed. This metric concerns both breadth and depth: a complete response should address all major sub-tasks, intermediate steps, and boundary conditions necessary to achieve the requested objective, while also furnishing enough textual and visual information for the user to understand and potentially reproduce the described process. High Completeness entails that no critical stages are omitted, that important concepts are not merely mentioned but adequately explained, and that images are supplied where they substantially enhance understanding of complex or abstract content. Outputs are penalized when they skip key steps, provide only superficial treatment of central issues, or fail to include visual support for pivotal multimodal tasks. A highly complete response therefore anticipates the user’s informational needs, providing a comprehensive and well-structured answer that is both conceptually and practically useful.

\textbf{Accuracy.}
To evaluate the model’s ability to follow precise instructions, we explicitly specify the required number of images to be produced in certain user prompts for interleaved text–image generation. The overall distribution of the requested image counts per prompt is shown in Fig.~\ref{fig:benchmark_stat} (c) in the main paper. By comparing the number of generated images with the required targets, we compute an accuracy metric that reflects the model’s fidelity to user-specified structural constraints. This metric captures the model’s robustness in adhering to quantitative generation requirements.

\begin{table*}[!t]
\centering
\caption{Comparison across single-modality generation tasks (understanding, image generation, and editing). $^{\dagger}$: Our in-house base model with thinking mode (enabled only for understanding). $^{\ddagger}$: Methods are using LLM rewriter.}
\label{tab:single_comparison_supp}
\begin{tabular}{l cc cc cc}
\toprule
\multirow{2}{*}{Model} &
\multicolumn{2}{c}{\textbf{Understanding}} &
\multicolumn{2}{c}{\textbf{Image Generation}} &
\multicolumn{2}{c}{\textbf{Image Editing}} \\
\cmidrule(lr){2-3} \cmidrule(lr){4-5} \cmidrule(lr){6-7}
& MMMU & MathVista & GenEval & DPG & ImgEdit & GEdit--EN \\
\midrule
\textbf{Visual Understanding Models} & & & & & & \\
LLaVA-1.5~\cite{liu2023visual} &67.8 & -- & --   & --    & --   & --   \\
LLaVA-NeXT~\cite{liu2024llavanext} & 51.1 &39.6  & --   & --    & --   & --   \\
InternVL3-38B~\cite{zhu2025internvl3}          & 69.7 & 76.3 & --   & --    & --   & --   \\
Ovis2-34B~\cite{lu2025ovis2}              & 66.7 & 76.1 & --   & --    & --   & --   \\
Qwen2.5-VL-32B$^{\dagger}$         & \best{75.1} & \best{84.7} & --   & --    & --   & --   \\
\midrule
\textbf{Text-to-Image Generation Models} & & & & & & \\
FLUX.1-dev~\cite{labs2025flux}             & --   & --   & 0.66 & 84.0  & --   & --   \\
SDXL~\cite{podell2023sdxl} & --   & -- & 0.55 & 74.7 &-- &-- \\
SD3-medium~\cite{esser2024scaling} & -- & -- & 0.62 & 84.1 & -- & -- \\
\midrule
\textbf{Image Editing Models} & & & & & & \\
Instruct-P2P~\cite{brooks2023instructpix2pix} & -- & -- & -- & -- & 1.88 & 3.68 \\
MagicBrush~\cite{zhang2023magicbrush} & -- & -- & -- & -- & 1.90 & 1.86 \\
AnyEdit~\cite{yu2025anyedit} & -- & -- & -- & -- & 2.45 & 3.21 \\
Step1X-Edit~\cite{liu2025step1x}            & --   & --   & --   & --    & 3.06 & \seco{6.70} \\
IC-Edit~\cite{zhang2025context} & -- & -- & -- & -- & 3.05 & 4.84 \\
\midrule
\textbf{Unified Models} & & & & & & \\
Janus-Pro~\cite{chen2025janus} & 41.0 & 42.5 &0.80 & 84.19 &-- &-- \\
Emu3~\cite{wang2024emu3} & 31.6 & -- & 0.66$^{\ddagger}$ & 80.60 & -- & -- \\
UniPic~\cite{wang2025skywork} & -- & -- & 0.86 & 85.50 & \seco{3.49} & 5.83 \\
MetaQuery-XL~\cite{pan2025transfer} & 58.6 &-- & 0.80$^{\ddagger}$ & 82.05 & -- & -- \\
Show-o2~\cite{xie2025show} & 48.9 &-- & 0.76 & \seco{86.14} & -- & -- \\
OmniGen~\cite{xiao2025omnigen} & -- & -- & 0.68 & 81.16 & 2.96 & 5.06 \\
OmniGen2~\cite{wu2025omnigen2} & 53.1 & --& 0.86$^{\ddagger}$ & 83.57 & 3.44 & 6.42 \\
BLIP3-o~\cite{chen2025blip3} & 58.6 & -- & 0.84$^{\ddagger}$ & 81.60 & -- & -- \\
Bagel~\cite{deng2025emerging}                  & 55.3 & 73.1 & \seco{0.88}$^{\ddagger}$ & 85.07 & 3.20 & 6.52 \\
UniWorld-V1~\cite{lin2025uniworld}            & 58.6 & --   & 0.84$^{\ddagger}$ & 81.38 & 3.26 & 4.85 \\
\rowcolor{Gray}
Wan-Weaver (Ours)                    & \seco{74.9} & \seco{84.3} & \best{0.89} & \best{87.21} & \best{4.31} & \best{7.39} \\
\bottomrule
\end{tabular}
\end{table*}
\subsection{Prompt for GPT-based Scoring}
The system prompt designed for GPT-based evaluator is presented in Table~\ref{tab:prompt_edit1},~\ref{tab:prompt_edit2},~\ref{tab:prompt_edit3}, and~\ref{tab:prompt_edit4}. The prompt instructs GPT to evaluate the generated interleaved text-image content based on the key evaluation metrics mentioned in Sec.~\ref{subsec:supp:evaluation_metrics} (excluding `Accuracy'). For each metric, the GPT-based evaluator is required to rate the results on a scale from 0 to 10. Importantly, unlike prior benchmarks that rely on GPT-based evaluators~\cite{zhou2025opening}, we provide explicit and fine-grained criteria for every integer score from 0 to 10. This design offers more precise guidance for the evaluator model and mitigates ambiguity during the scoring process.

\begin{figure*}[!ht]
    \centering
    \includegraphics[width=0.75\linewidth]{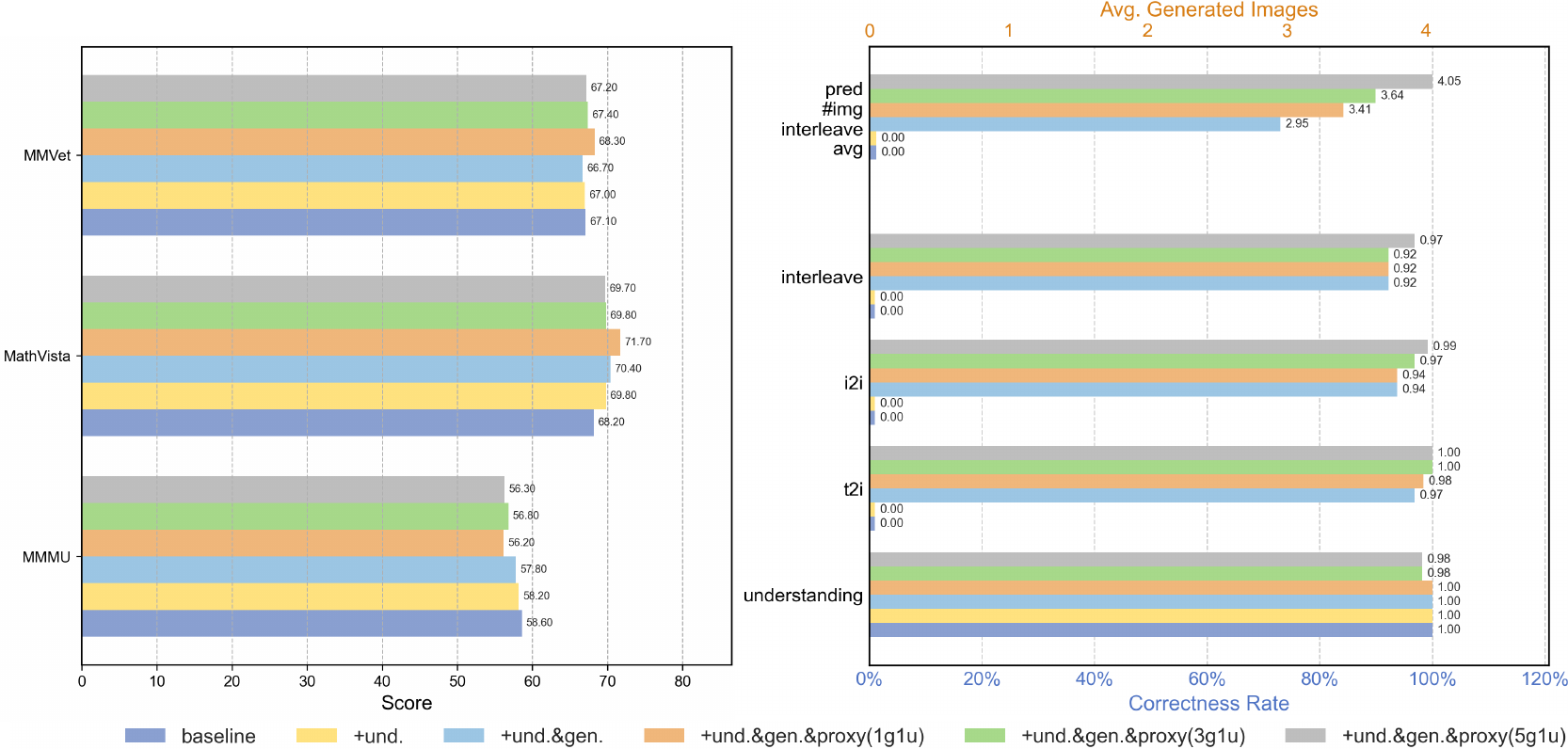} 
    \caption{Impact of training on different types of data on our planner. (left) Performance on understanding metrics. (right) Token prediction statistics per task.}
    \label{fig:mmu_stat_supp}
\end{figure*}

\subsection{Examples}
\begin{figure*}[!ht]
    \centering
    \includegraphics[width=1\linewidth]{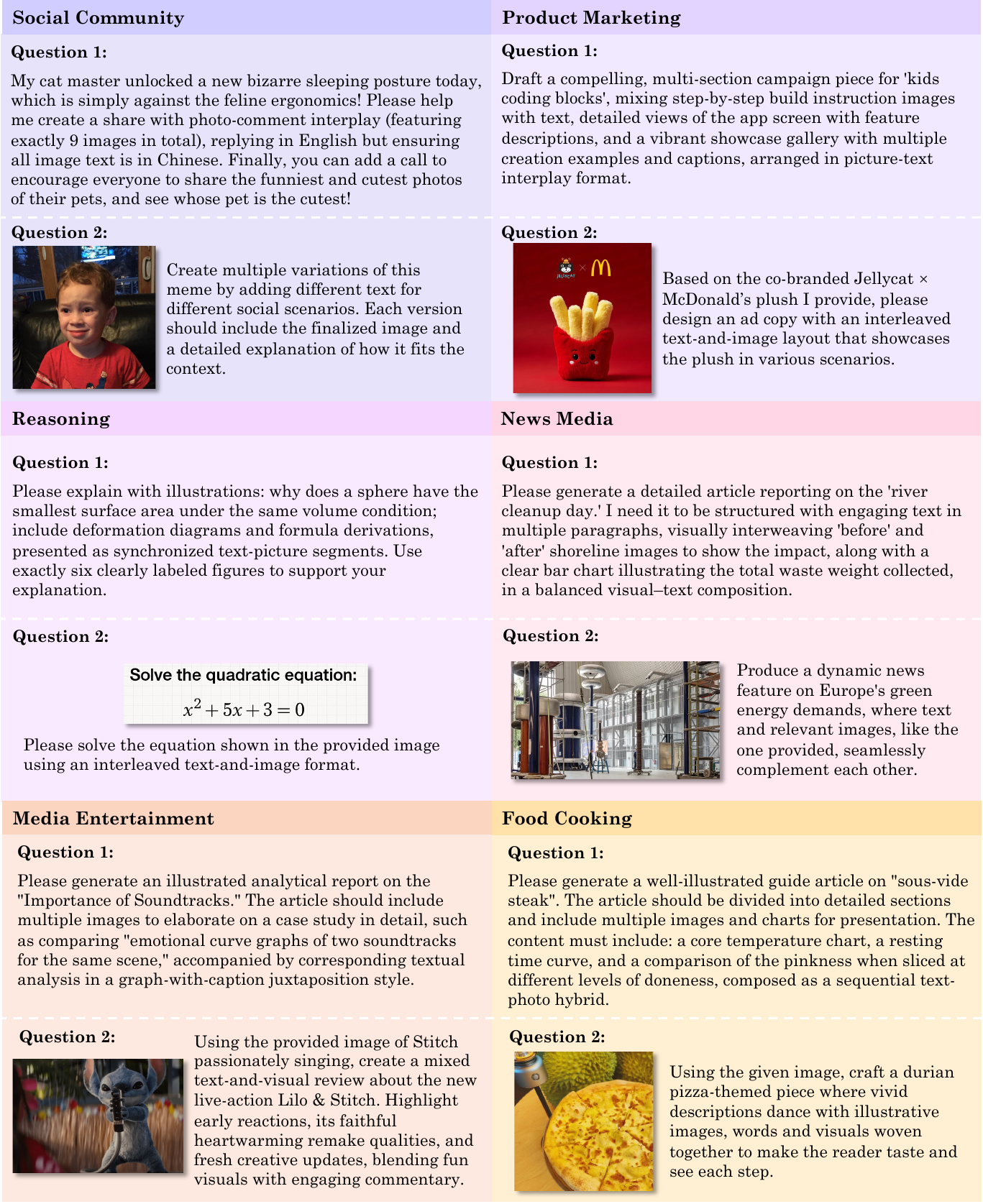} 
    \vspace{-1em}
    \caption{Examples of test samples from WeaverBench. For each category, we show one text-only user prompt and one mixed text–image user prompt (Part I).}
    \label{fig:benchmark_sample1}
\end{figure*}

\begin{figure*}[!ht]
    \centering
    \includegraphics[width=1\linewidth]{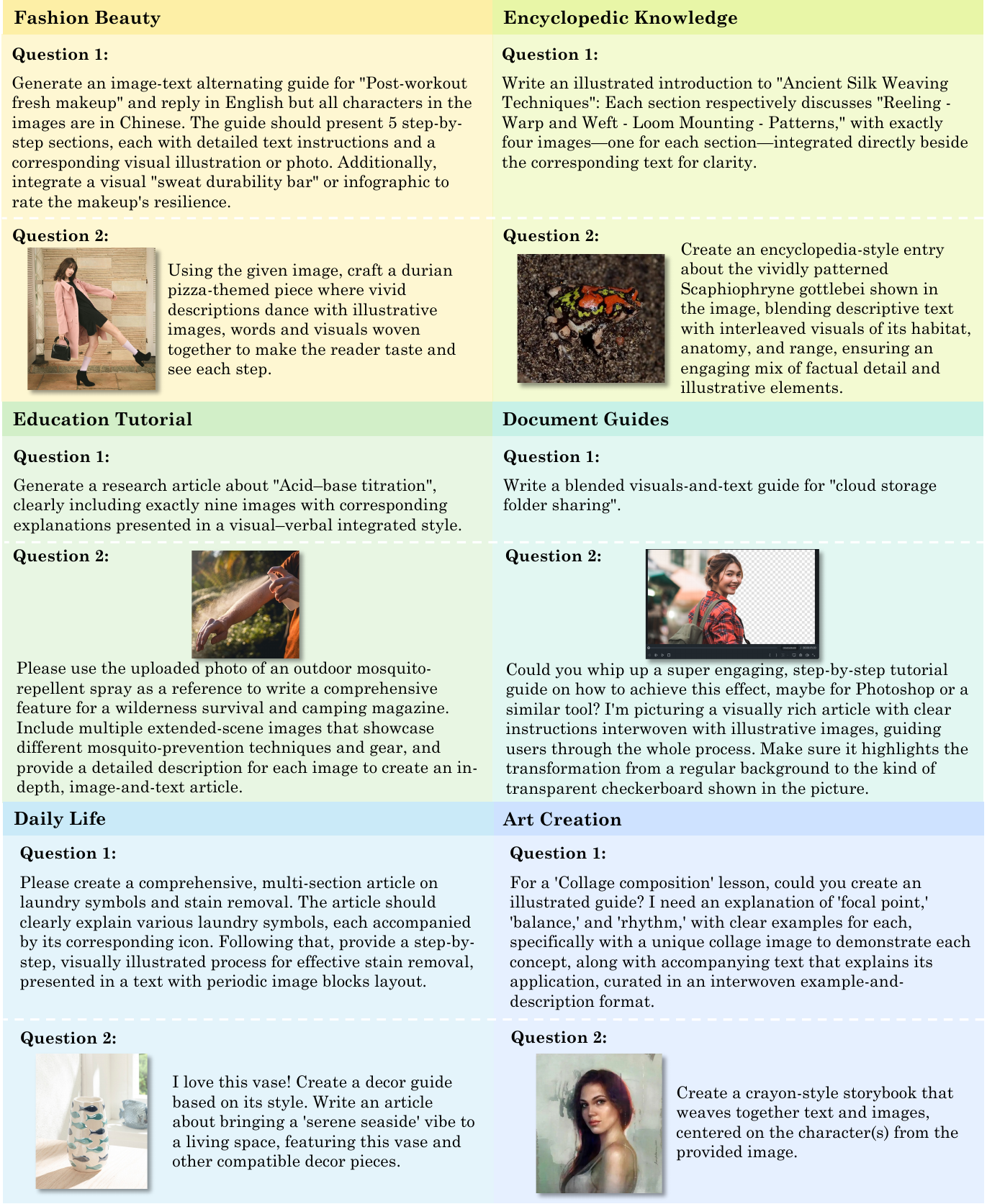} 
    \vspace{-1em}
    \caption{Examples of test samples from WeaverBench. For each category, we show one text-only user prompt and one mixed text–image user prompt (Part II).}
    \label{fig:benchmark_sample2}
\end{figure*}

\begin{figure*}[!ht]
    \centering
    \includegraphics[width=1\linewidth]{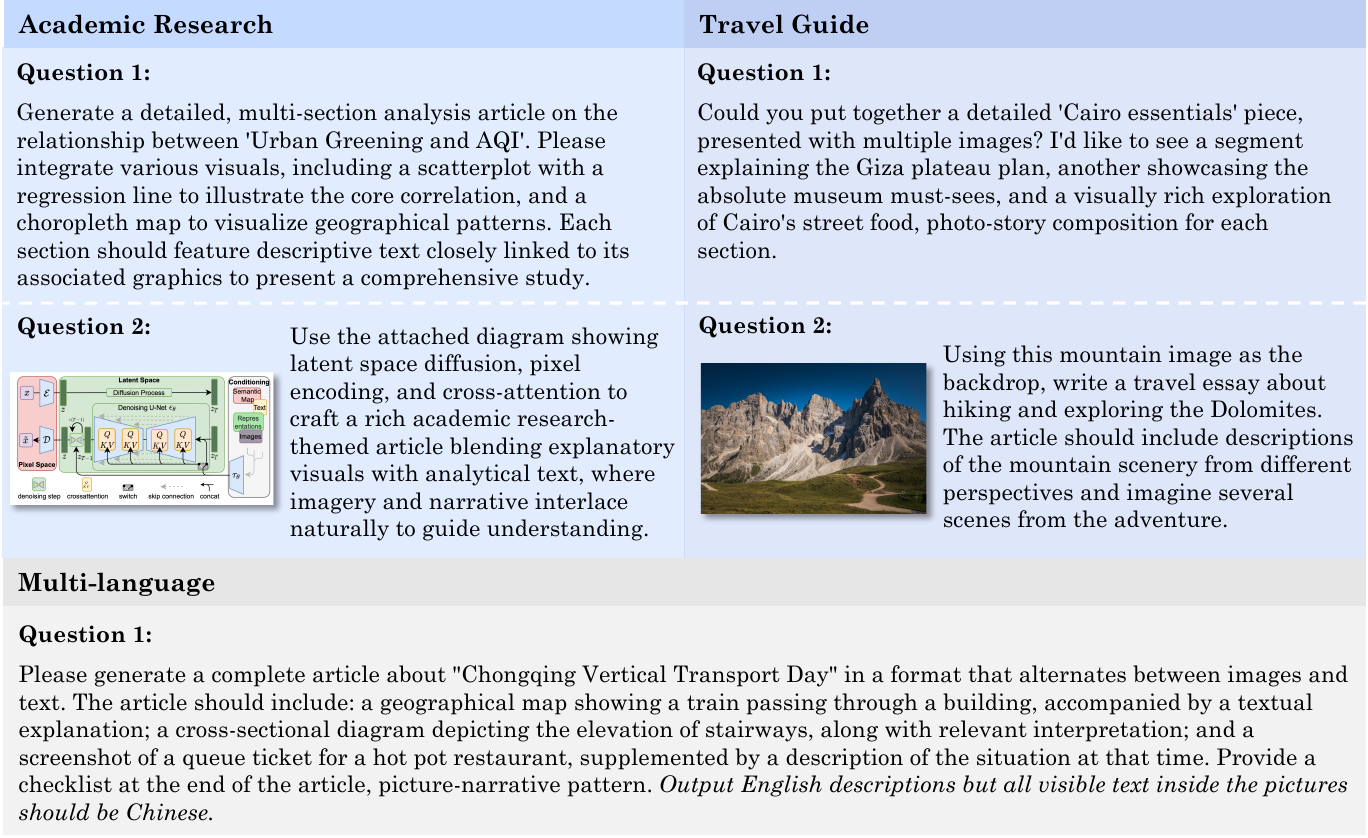} 
    \vspace{-1em}
    \caption{Examples of test samples from WeaverBench. For each category, we show one text-only user prompt and one mixed text–image user prompt (Part III).}
    \label{fig:benchmark_sample3}
\end{figure*}

Illustrations of the representative examples of the 15 categories are provided in Fig.~\ref{fig:benchmark_sample1},~\ref{fig:benchmark_sample2}, and~\ref{fig:benchmark_sample3}, highlighting the breadth and diversity of testing scenarios covered by WeaverBench. For each category, we present both text-only user prompts and mixed text–image prompts to reflect different forms of real-world inputs.

\begin{figure*}[t]
    \centering
    \includegraphics[width=1\linewidth]{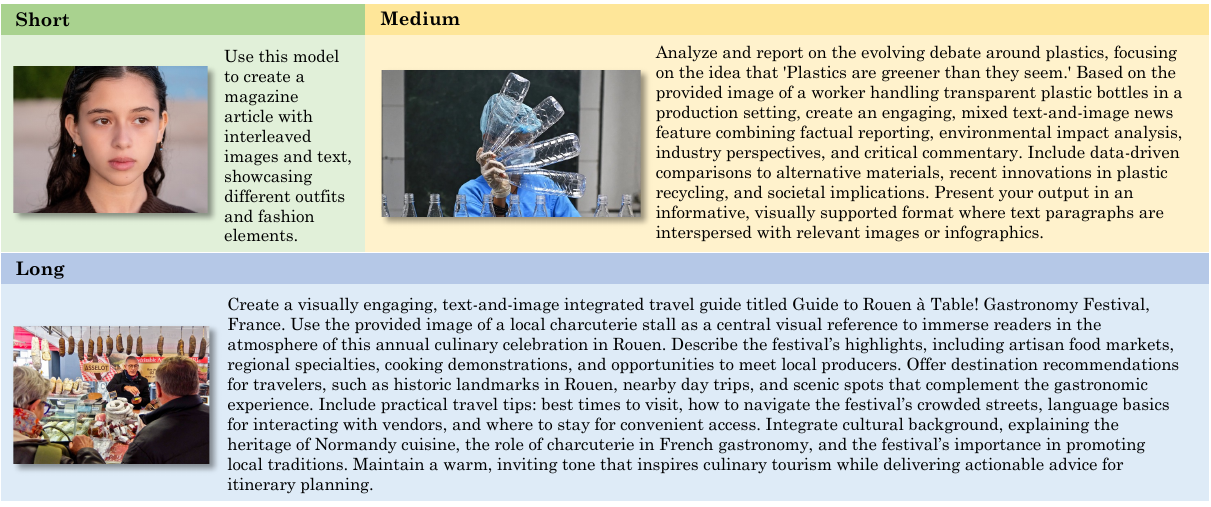} 
    \caption{Examples of short, medium, and long user prompts in WeaverBench.}
    \label{fig:benchmark_sml}
\end{figure*}

To further emulate the practical usage patterns, where users may issue instructions with varying levels of specificity, we also construct prompts of different lengths and detail levels. The overall distribution of prompt lengths is reported in Fig.~\ref{fig:benchmark_stat} (b) in the main paper, with concrete examples provided in Fig.~\ref{fig:benchmark_sml}.

\begin{figure*}[t]
    \centering
    \includegraphics[width=0.9\linewidth]{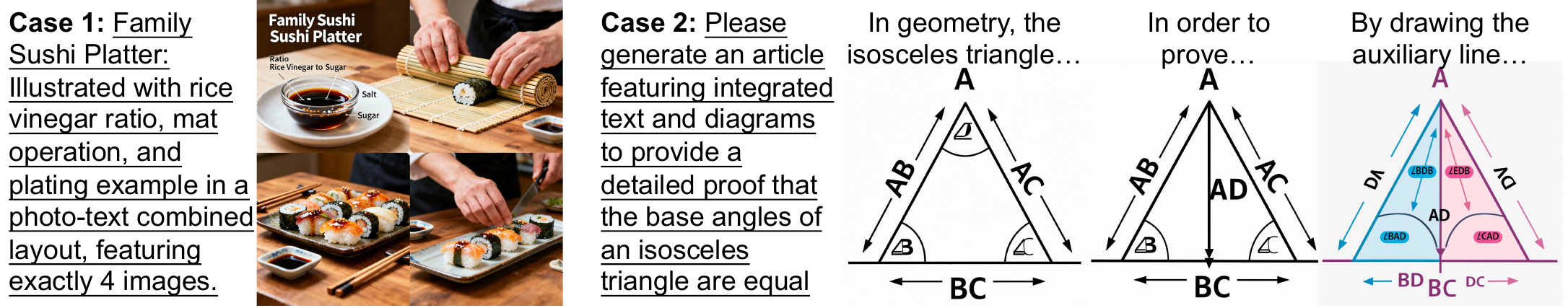} 
    \caption{Visual examples of failure cases.}
    \label{fig:failure_case}
\end{figure*}

\begin{figure*}[!ht]
    \centering
    \vspace{-2.5em}
    \includegraphics[width=1\linewidth]{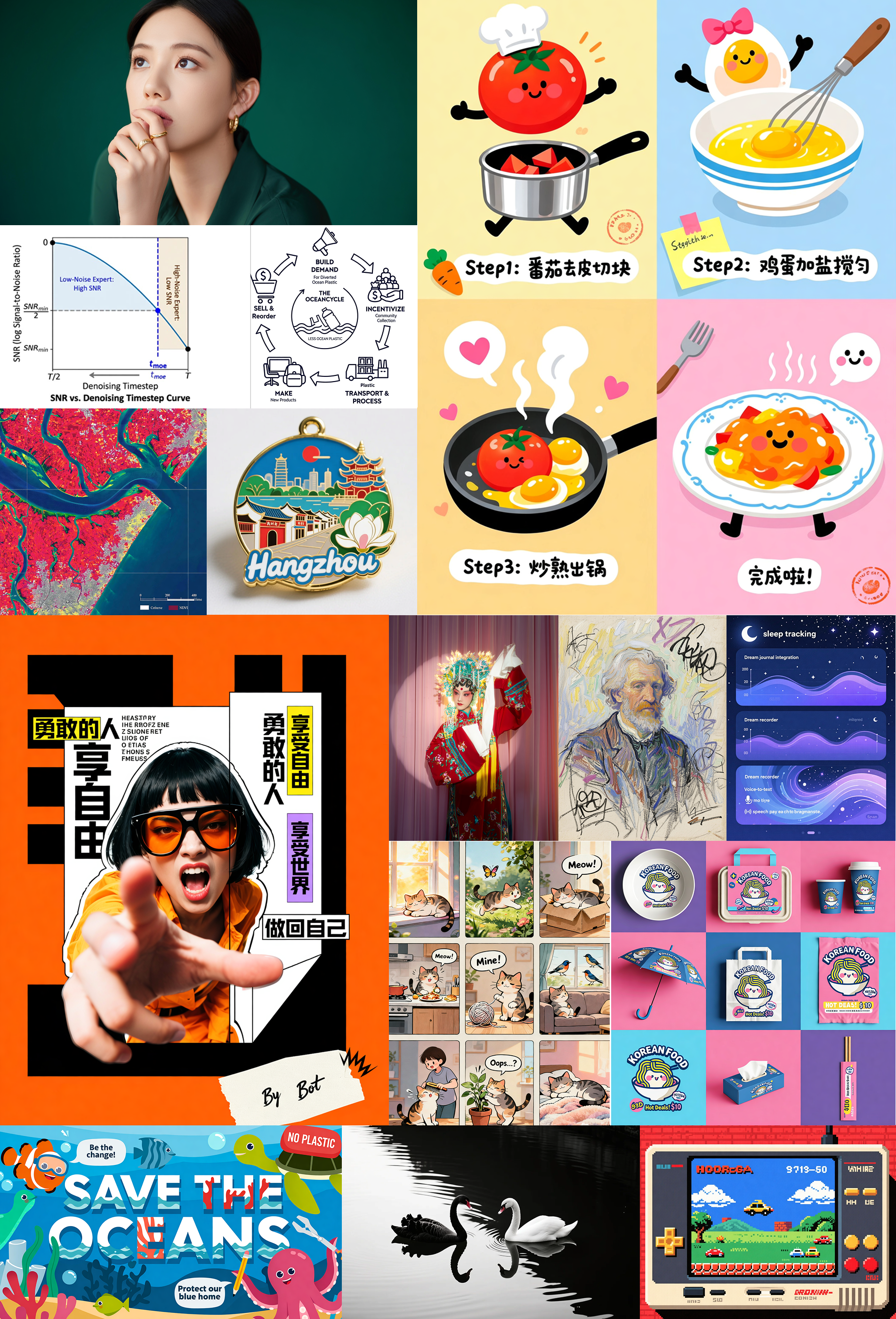} 
    \vspace{-2em}
    \caption{Showcase of Wan-Weaver in general text-to-image generation.}
    \label{fig:t2i_supp}
\end{figure*}
\begin{figure*}[t]
    \centering
    \vspace{-2em}
    \includegraphics[width=1\linewidth]{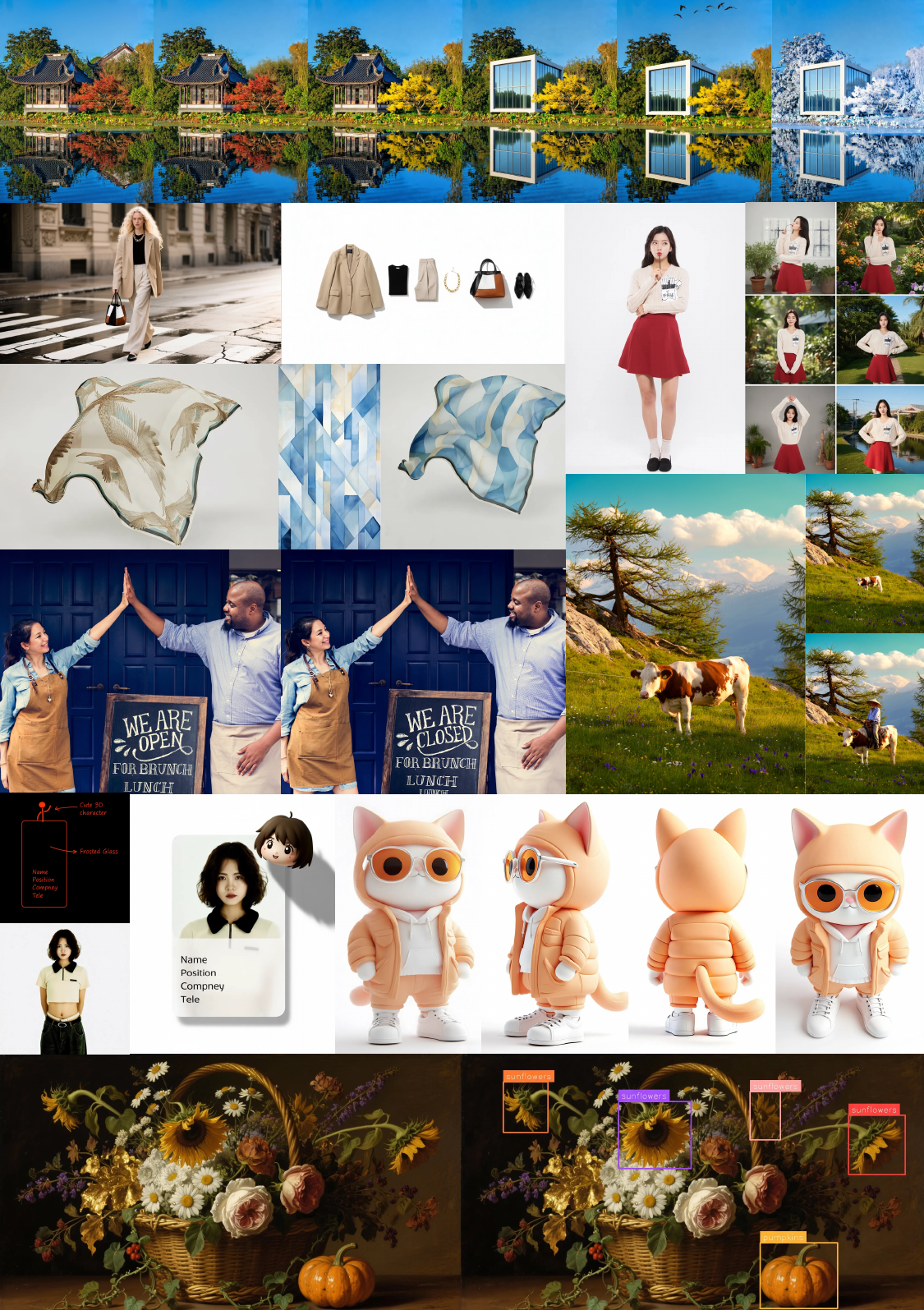} 
    \vspace{-2em}
    \caption{Showcase of Wan-Weaver in general image-to-image generation, including object addition, removal, replacement, element extraction, pose manipulation, text editing, style/texture transfer, novel view synthesis, object detection/segmentation, and more.}
    \label{fig:i2i_supp}
\end{figure*}
\begin{figure*}[t]
    \centering
    \vspace{-2.5em}
    \includegraphics[width=0.95\linewidth]{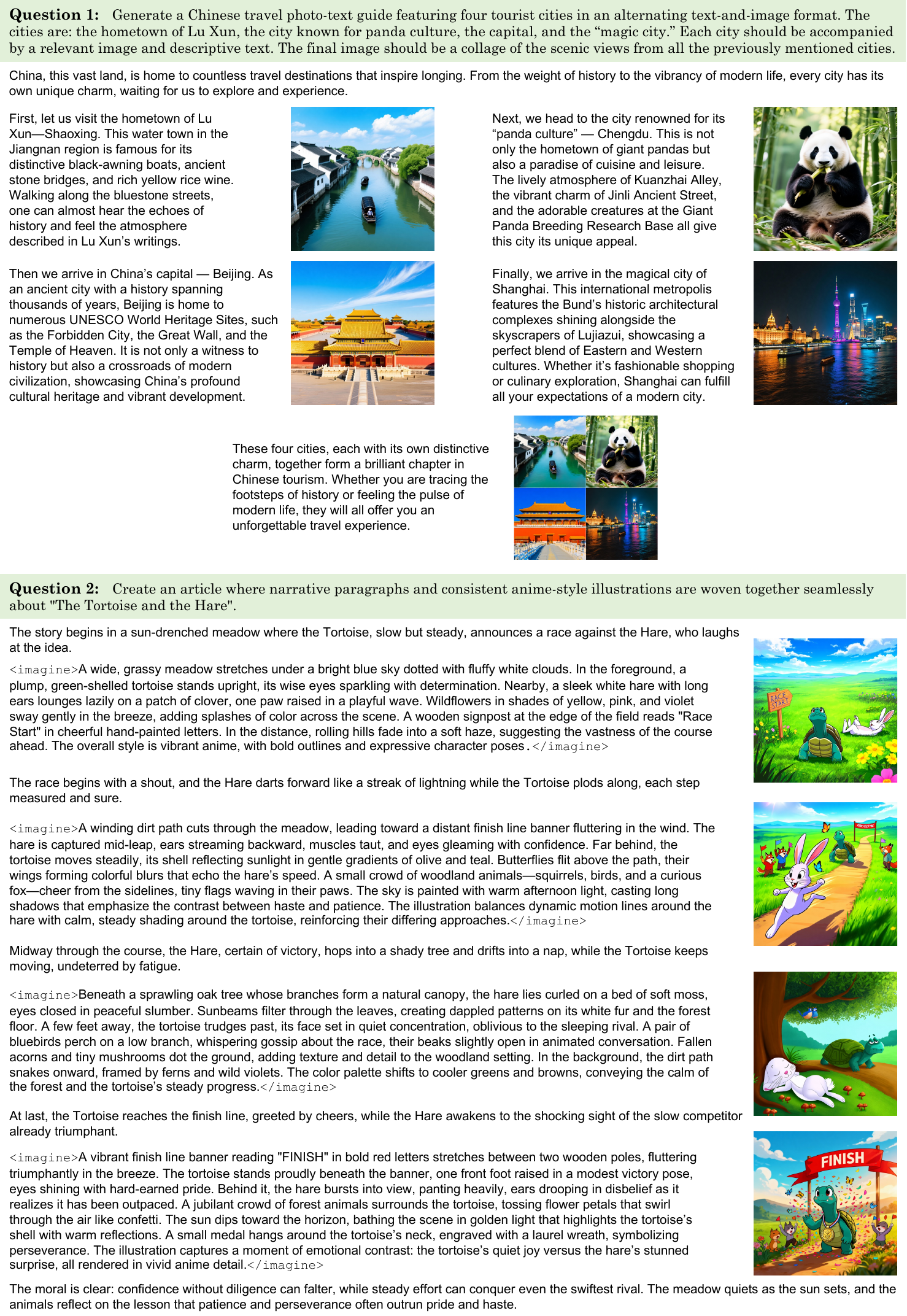} 
    \vspace{-1em}
    \caption{Showcase of Wan-Weaver in interleaved text-image generation.}
    \label{fig:interleave_supp}
\end{figure*}

\section{Additional Analysis}
\paragraph{Impact of Fine-tuned Planner on Visualizer.} Since our framework adopts a decoupled training strategy, the planner and visualizer are optimized in separate stages. This raises a natural question: after we initialize the planner from a pre-trained VLM and train the visualizer accordingly, will subsequent fine-tuning of the planner for planning capabilities introduce feature shifts that negatively affect image generation?
To examine this, we fine-tune the visualizer again using the planner after its planning-oriented fine-tuning, and compare the resulting vision loss with that of the original visualizer trained with the pre-trained planner. As shown in Fig.~\ref{fig:vision_loss} in the main paper, \ie, V (T2I+SI2I+MI2I) vs. aP V (T2I+SI2I+MI2I), the two loss curves are extremely close. This indicates that the planner's features change only minimally during planning fine-tuning and do not adversely impact the visualizer's generation quality.

\paragraph{Feature Modeling in Planner.}
Due to the limited space in the main paper, we provide a more detailed analysis here. To investigate how acquiring various levels of planning ability affects the underlying understanding capability, we perform planner tuning on different data compositions. The baseline corresponds to the VLM-initialized planner without any planning ability. As illustrated in Fig.~\ref{fig:mmu_stat_supp}, we construct several variants trained with different subsets of data: (1) \textbf{+und.\&gen.\&proxy}, which endows the planner with comprehensive planning skills across all tasks, including the generation of dense prompts for interleaved outputs; (2) \textbf{+und.\&gen.}, which preserves basic understanding while introducing generation-oriented planning; and (3) \textbf{+und.}, which preserves only the original understanding capability.

From Fig.~\ref{fig:mmu_stat_supp} (left), we observe that the understanding performance remains highly stable across all configurations, indicating that introducing planning ability does not compromise the model’s core understanding competency. Beyond understanding accuracy, we further examine whether multi-task planning enables the planner to produce task-appropriate modality-specific patterns. Specifically, we compute the token-prediction accuracy of structural plan tokens. Given a user prompt, the planner must correctly produce the modality pattern, \eg, no \texttt{<BOI>} token for understanding tasks, exactly one for T2I/I2I tasks, and at least one for interleaved generation. As shown in Fig.~\ref{fig:mmu_stat_supp} (right), we find that without any generation data (gen.), the model fails to emit image-generation signals. As the proportion of generation-oriented data increases, from zero gen. data, to 1g1u (1:1 ratio of generation to understanding), 3g1u, and 5g1u, the planner's proficiency in generative planning improves substantially. For interleaved tasks in particular, the average number of predicted image starting tokens gradually increases as well.
Balancing the overall planning reliability and understanding stability, we adopt the 5g1u ratio as the final data composition for planner tuning.

\section{More Quantitative Results}

\paragraph{Single Modality Generation.} In addition to the quantitative results reported in Table~\ref{tab:single_comparison} of the main paper, we present additional baseline performance in Table~\ref{tab:single_comparison_supp}. The conclusions remain consistent: our method demonstrates superior performance compared with previous unified and specialized generation models. 

\paragraph{Detailed Results on WeaverBench.} We provide more fine-grained results of model performance on WeaverBench, serving as a complement to Table~\ref{tab:weaver_bench_results} in the main paper. Note that the CC and IC scores reported in the main paper correspond to the averages of their respective subcomponents. In addition to the methods included previously, we further report the results of Anole~\cite{chern2024anole} and Emu3.5~\cite{cui2025emu35}, representative models with general interleaved-generation capabilities, to offer a more comprehensive comparison. Note Emu3.5 technical report was just released on Oct.30, 2025, which is concurrent with our work.
Moreover, we introduce an additional metric: Acc (accuracy), which measures whether the model generates the exact number of images specified by the user. Experimental results indicate that our approach follows user instructions much more reliably, achieving significantly higher accuracy than all competing baselines.

\input{supp/tables/win_rate}
\paragraph{Win Rate Comparison on OpenING.} In addition to the GPT-scoring results shown in Table~1, we also report the win rate comparison on OpenING~\cite{zhou2025opening}, which is tabulated in Table~\ref{tab:winrate_supp}. Our model consistently achieves top-tier performance across all GPT-based evaluation metrics, surpassing all competing systems by a clear margin. It delivers the highest FDT accuracy and remains robust under different tie-breaking strategies, indicating strong pairwise preference alignment and stable comparative quality. Under IntJudge evaluation, our model maintains competitive rankings and demonstrates notably strong performance in tie-adjusted settings, highlighting its reliability and consistency across different evaluators. Overall, the results show that our method achieves state-of-the-art preference quality and general robustness in interleaved text-image generation.

\section{More Qualitative Results}
\label{sec:more_qualitative}
\subsection{Text-to-Image Generation}
As illustrated in Fig.~\ref{fig:t2i_supp}, our model supports a broad spectrum of text-to-image generation capabilities, including varied aspect ratios, multiple visual styles, accurate text rendering, comic and panel layouts, charts, flow diagrams, posters, and more.

\subsection{Text-Image-to-Image Generation}
Fig.~\ref{fig:i2i_supp} presents a broad range of Wan-Weaver’s text-image-to-image generation capabilities, covering object addition, deletion, replacement, element extraction, pose manipulation, text modification, style and texture transfer, novel-view synthesis, object detection and segmentation, and more.

\subsection{Interleaved Text-image Generation}
Additional interleaved text–image generation examples are shown in Fig.~\ref{fig:interleave_supp}. The results demonstrate that our method is capable of generating text and images that are contextually coherent. Our method also exhibits notable reasoning abilities: although the user prompt does not directly specify `Shanghai', but instead refers to it as the `magic city' in China, the model is nevertheless able to produce accurate textual descriptions and corresponding images. The final image represents a more challenging case, as the model must infer which references to use and what content should be generated. Our results show that the model correctly incorporates information from all four preceding images and produces a coherent collection while preserving the detailed characteristics of the original visuals. The second case in Fig.~\ref{fig:interleave_supp} shows the dense prompt generated within the \texttt{<imagine>}\texttt{</imagine>} tags.
We also provide further qualitative comparisons in an accompanying local HTML file for detailed inspection.

\section{Limitations}
\label{sec:limitations}
Despite the strong performance of our unified model for interleaved multi-modal generation, several limitations remain.

First, the current model requires users to specify a fixed resolution and aspect ratio for all generated images, or to manually predefine these settings for each image in the sequence. Such constraints limit flexibility and are not aligned with real-world usage, where image composition, aspect ratio, and level of detail should ideally be determined automatically based on the evolving narrative and visual semantics. A more user-friendly design would involve adaptive resolution planning, allowing the model to infer the appropriate size and aspect ratio from the content itself, supporting richer visual–textual storytelling and more practical downstream applications.

Second, our planner–visualizer framework inherently follows a general sequential generation process, where all previously generated content, \ie, both text and images, is fed back as reference information to guide the next-step generation. Although the visualizer's per-image denoising can be parallelized across multiple GPUs, the overall workflow remains sequential, and the amount of conditioning grows with each newly generated image. As a result, generation efficiency progressively decreases, and GPU memory consumption increases over time. This sequential dependency poses a bottleneck for long interleaved sequences, suggesting the need for more efficient mechanisms to reduce accumulated conditioning while preserving contextual fidelity.

Third, although the planner–visualizer design demonstrates that understanding can facilitate better generation to some extent, the reverse effect is limited: enhanced generative ability does not meaningfully improve the model's understanding capability. Achieving such bidirectional reinforcement likely requires more specialized training strategies, tailored multi-modal datasets, or unified representation mechanisms that tightly couple understanding and generation. We leave these as future works.

In addition, we show more failure cases in Fig.~\ref{fig:failure_case} regarding structural layout adherence and complex geometry reasoning: 
1) Structural Collapse into Grid Layout.
A primary limitation involves the model's occasional failure to maintain the requested interleaved document structure. As shown in Case 1 of Fig.~\ref{fig:failure_case}, when prompted to generate a photo-text combined layout with multiple images, the model tends to collapse these elements into a single-canvas grid layout rather than distributing them as separate entities within the text stream. This `grid-image bias' suggests that the model sometimes conflates the concept of `multiple images' with a `composite collage', revealing a deficiency in its ability to orchestrate fine-grained document-level formatting and sequential image placement.
2) Logical Consistency and Geometric Reasoning.
Case 2 highlights a limitation in handling high-level abstract reasoning and symbolic grounding. Although the model attempts to generate diagrams for a geometry proof, it fails to maintain logical consistency between the visual auxiliary lines and the corresponding textual labels. The rendered text within the images often suffers from hallucinations or symbolic confusion (e.g., misaligned vertex labels and inconsistent geometric topology).

\begin{table*}[!ht]\centering
\begin{minipage}{0.9\textwidth}\vspace{0mm}    
    \centering
    \begin{tcolorbox} 
        \centering
        \hspace{-6mm}
        \begin{tabular}{p{0.9\textwidth}}
        \hspace{1mm}
        \begin{minipage}{0.9\textwidth}
        \texttt{\#\#\#[System Prompt]}\\
You are an expert in multimodal content evaluation. Your task is to assess interleaved image-text content generated by a model.\\
The model’s input starts with "INPUT:" and can be a question or instruction requesting such content.\\
The model’s output starts with "OUTPUT:" and may contain interleaved images and texts based on the input.\\
Image Representation:\\
- The image for step i will be enclosed by \verb|<IMG_i> and </IMG_i>|.\\
- Images are numbered sequentially from 0 to N (including images in the input).\\
- You can directly accept image inputs for this evaluation.\\
---\\
Evaluation Criteria – Main Categories and Subcategories\\
1. \textbf{Prompt Adherence}: Evaluates whether the combined image-text output directly addresses the user’s request, fulfills the specified task, and provides a correct or useful solution. Penalize outputs that are off-topic, incomplete with respect to the core task, or fail to respond to key aspects of the instruction.\\
\textit{Scoring Guidelines:\\
- 0: Prompt Adherence—Output is missing, irrelevant, or ignores the request entirely.\\
- 1: Prompt Adherence—Barely related; no real attempt to address the task.\\
- 2: Prompt Adherence—Addresses the wrong task or gives fundamentally incorrect content.\\
- 3: Prompt Adherence—Covers only a minor part; omits essential requirements.\\
- 4: Prompt Adherence—Partially correct but misses critical components.\\
- 5: Prompt Adherence—Meets basic intent but has noticeable omissions or errors.\\
- 6: Prompt Adherence—Mostly complete with minor flaws; solution is usable.\\
- 7: Prompt Adherence—Accurate, complete, and directly addresses the request.\\
- 8: Prompt Adherence—Fully fulfills task with relevant, useful details.\\
- 9: Prompt Adherence—Exceeds expectations by covering implicit needs precisely.\\
- 10: Prompt Adherence—Perfect execution with complete fidelity to the request.\\}
\\
2. \textbf{Narrative Coordination}: Assesses whether images and text are placed at appropriate points in the response (e.g., image appears when a visual illustration is needed), and whether their sequencing supports clear understanding of the multi-step process or narrative.\\
\textit{Scoring Guidelines:\\
  - 0: Narrative Coordination—No structure; media and text are random or single-modality.\\
  - 1: Narrative Coordination—No intentional sequencing; placement appears arbitrary.\\
  - 2: Narrative Coordination—Sequence is chaotic; visuals disrupt rather than support understanding.\\
  - 3: Narrative Coordination—Key steps lack needed visuals or textual context.\\
  - 4: Narrative Coordination—Poor timing or placement creates confusion or gaps.\\
  - 5: Narrative Coordination—Inconsistent flow; some sections poorly organized.\\
  - 6: Narrative Coordination—Generally logical; small sequencing issues don’t hinder understanding.\\
  - 7: Narrative Coordination—Images appear where helpful; text and visuals align clearly.\\
  - 8: Narrative Coordination—Thoughtful sequencing; media and text enhance each other seamlessly.\\
  - 9: Narrative Coordination—Pedagogically sound structure that anticipates user comprehension.\\
  - 10: Narrative Coordination—Elegant, flawless integration of text and images for maximum clarity.}
          \end{minipage}
        \end{tabular}
    \end{tcolorbox}
    \vspace{-2mm}
    \caption{The system prompts for obtaining detailed scores from GPT-based evaluators (Part I).}
    \label{tab:prompt_edit1}
    \end{minipage}
    \vspace{-2mm}
\end{table*}

\begin{table*}[!ht]\centering
\begin{minipage}{0.9\textwidth}\vspace{0mm}    
    \centering
    \begin{tcolorbox} 
        \centering
        \hspace{-6mm}
        \begin{tabular}{p{0.9\textwidth}}
        \hspace{1mm}
        \begin{minipage}{0.9\textwidth}
3. \textbf{Content Consistency}\\
  1). Intra-output Coherence: Evaluates whether each generated image matches the semantic content, tone, and stylistic description of its accompanying or adjacent generated text (e.g., if the text says “a watercolor sketch of a rainy street,” the image should reflect that medium and mood).\\
  2). Input-to-Image Fidelity: Assesses whether generated images respect explicit or implicit visual constraints from the input (e.g., character descriptions, setting, or artistic style mentioned in the prompt).\\
  3). Input-to-Text Fidelity: Evaluates whether the generated text accurately reflects the intent, domain, tone, and factual expectations set by the input (e.g., a request for a scientific explanation should not yield poetic prose unless instructed).\\
\textit{Scoring Guidelines:\\
  - 0: Intra-output Coherence — Image directly contradicts its adjacent text; Input-to-Image Fidelity — Ignores all visual constraints from the input; Input-to-Text Fidelity — Text disregards input intent, domain, or task.\\
  - 1: Intra-output Coherence — Image and text appear randomly paired; Input-to-Image Fidelity — Violates core visual details like character or setting; Input-to-Text Fidelity — Uses wrong tone, genre, or style entirely.\\
  - 2: Intra-output Coherence — Frequent mismatches in content, mood, or style; Input-to-Image Fidelity — Misses key visual specifications from the prompt; Input-to-Text Fidelity — Misunderstands the purpose or factual expectations.\\
  - 3: Intra-output Coherence — Rarely aligned; mostly inconsistent pairings; Input-to-Image Fidelity — Partially follows input but omits essential visual cues; Input-to-Text Fidelity — Occasionally relevant but largely off-topic or inaccurate.\\
  - 4: Intra-output Coherence — Matches only basic or literal descriptions; Input-to-Image Fidelity — Honors at most one visual constraint (e.g., color or style); Input-to-Text Fidelity — Weakly aligned in isolated phrases only.\\
  - 5: Intra-output Coherence — Moderate alignment in some image-text pairs; Input-to-Image Fidelity — Roughly respects general style or setting; Input-to-Text Fidelity — Generally on-topic but inconsistent in tone or accuracy.\\
  - 6: Intra-output Coherence — Minor drifts in mood, medium, or detail; Input-to-Image Fidelity — Mostly adheres to explicit and implicit visual guidance; Input-to-Text Fidelity — Accurately reflects intent with small lapses.\\
  - 7: Intra-output Coherence — Strong match in semantics, mood, and style; Input-to-Image Fidelity — Faithfully respects all key visual constraints; Input-to-Text Fidelity — Correctly captures domain, tone, and factual expectations.\\
  - 8: Intra-output Coherence — Seamless harmony between image and its text; Input-to-Image Fidelity — Precisely fulfills all stated and implied visual requirements; Input-to-Text Fidelity — Text perfectly matches input instructions and context.\\
  - 9: Intra-output Coherence — Near-perfect unity in meaning and aesthetic; Input-to-Image Fidelity — Fully honors nuanced visual and contextual cues; Input-to-Text Fidelity — Demonstrates deep, accurate understanding of user intent.\\
  - 10: Intra-output Coherence — Image and text feel indistinguishably unified; Input-to-Image Fidelity — Flawless execution of every visual specification; Input-to-Text Fidelity — Perfectly embodies the input’s intent, tone, and detail as if user-authored.\\}
  \\
4. \textbf{Image Consistency}\\
  1). Entity Consistency: Checks that recurring objects, characters, or scenes maintain consistent appearance, attributes, and spatial relationships across all relevant images (e.g., a red backpack in step 1 should remain red and recognizable in step 3).
            \end{minipage}
        \end{tabular}
    \end{tcolorbox}
    \vspace{-2mm}
    \caption{The system prompts for obtaining detailed scores from GPT-based evaluators (Part II).}
    \label{tab:prompt_edit2}
    \end{minipage}
    \vspace{-2mm}
\end{table*}

\begin{table*}[!ht]\centering
\begin{minipage}{0.9\textwidth}\vspace{0mm}    
    \centering
    \begin{tcolorbox} 
        \centering
        \hspace{-6mm}
        \begin{tabular}{p{0.9\textwidth}}
        \hspace{1mm}
        \begin{minipage}{0.9\textwidth}
  2). Visual Style Uniformity: Evaluates consistency in artistic and photographic properties across all generated images, including media type (e.g., oil painting vs. photo), color palette, lighting, saturation, contrast, perspective, and overall aesthetic feel (e.g., consistently cartoonish or consistently photorealistic).\\
\textit{
Scoring Guidelines:\\
  - 0: Entity Consistency — Recurring entities vanish or change beyond recognition; Visual Style Uniformity — Style shifts randomly with no consistency across images.\\
  - 1: Entity Consistency — No effort to maintain identity; repeated items appear unrelated; Visual Style Uniformity — Each image uses a different medium, palette, or aesthetic.\\
  - 2: Entity Consistency — Drastic changes make entities impossible to track; Visual Style Uniformity — Erratic jumps between styles break visual coherence completely.\\
  - 3: Entity Consistency — Frequent unintended changes in color, shape, or pose; Visual Style Uniformity — Disruptive variations in lighting, perspective, or rendering type.\\
  - 4: Entity Consistency — Sometimes recognizable but often altered inconsistently; Visual Style Uniformity — Mixed aesthetics (e.g., photorealistic and cartoon) or unstable color grading.\\
  - 5: Entity Consistency — Core features partially retained but with noticeable drift; Visual Style Uniformity — General style trend present but with clear inconsistencies.\\
  - 6: Entity Consistency — Most recurring entities remain identifiable across steps; Visual Style Uniformity — Style largely uniform with only minor, non-disruptive variations.\\
  - 7: Entity Consistency — Appearance and spatial relationships stay consistent; Visual Style Uniformity — Cohesive look in medium, lighting, tone, and perspective throughout.\\
  - 8: Entity Consistency — Attributes reliably preserved for easy tracking over time; Visual Style Uniformity — Fully unified media type, palette, lighting, and composition.\\
  - 9: Entity Consistency — Precise visual identity maintained in every appearance; Visual Style Uniformity — Feels like frames from a single professional visual narrative.\\
  - 10: Entity Consistency — Flawless preservation of shape, color, pose, and context across all steps; Visual Style Uniformity — All images indistinguishable in style—perfect visual harmony.\\}
  \\
5. \textbf{Completeness}: Verifies that all expected steps, components, or stages implied or explicitly requested in the input are present in the output, with sufficient explanatory depth in text and appropriate visual support where needed. Penalize omissions, underspecification, or glossing over critical subtasks.\\
\textit{Scoring Guidelines:\\
  - 0: Completeness — Ignores all explicit and implicit expectations; provides no steps, explanation, or visual support.\\
  - 1: Completeness — Omits nearly all required content; lacks structure and meaningful text or images.\\
  - 2: Completeness — Misses critical stages; explanations are extremely shallow and key visuals are absent.\\
  - 3: Completeness — Skips major subtasks; lacks necessary detail and omits visuals for pivotal steps.\\
  - 4: Completeness — Misses important components; explanations are insufficient and core parts lack proper visual support.\\
  - 5: Completeness — Covers main points but glosses over several areas; critical subtasks receive only superficial treatment.\\
  - 6: Completeness — Includes all essential steps with minor omissions; text is sufficient and visuals mostly appropriate.}
            \end{minipage}
        \end{tabular}
    \end{tcolorbox}
    \vspace{-2mm}
    \caption{The system prompts for obtaining detailed scores from GPT-based evaluators (Part III).}
    \label{tab:prompt_edit3}
    \end{minipage}
    \vspace{-2mm}
\end{table*}

\begin{table*}[!ht]\centering
\begin{minipage}{0.9\textwidth}\vspace{0mm}    
    \centering
    \begin{tcolorbox} 
        \centering
        \hspace{-6mm}
        \begin{tabular}{p{0.9\textwidth}}
        \hspace{1mm}
        \begin{minipage}{0.9\textwidth}
\textit{
  - 7: Completeness — All major steps present with adequate depth; visuals provided where needed for clarity.\\
  - 8: Completeness — Rich explanations and high-quality, relevant visuals for all significant components.\\
  - 9: Completeness — Includes every expected element and anticipates nuanced needs with insightful augmentation.\\
  - 10: Completeness — Perfectly exhaustive; every requirement met with ideal detail and precisely matched visuals.\\}
---\\
Output JSON Structure:
\begin{small}
\begin{verbatim}
Only four main categories are shown; each "Score" is the average of 
its subcategory scores.
{
  "scores": {
    "Prompt Adherence": {
      "Score": 0-10,
      "Justification": "Brief explanation of any issues identified"
    },
    "Narrative Coordination": {
      "Score": 0-10,
      "Justification": "Brief explanation of any issues identified"
    },
    "Image Text Consistency (Intra-output Coherence)": {
      "Score": 0-10,
      "Justification": "Brief explanation of any issues identified"
    },
    "Image Text Consistency (Input-to-Image Fidelity)": {
      "Score": 0-10,
      "Justification": "Brief explanation of any issues identified"
    },
    "Image Text Consistency (Input-to-Text Fidelity)": {
      "Score": 0-10,
      "Justification": "Brief explanation of any issues identified"
    },
    "Multi step Consistency (Entity Consistency)": {
      "Score": 0-10,
      "Justification": "Brief explanation of any issues identified"
    },
    "Multi step Consistency (Visual Style Uniformity)": {
      "Score": 0-10,
      "Justification": "Brief explanation of any issues identified"
    },
    "Completeness": {
      "Score": 0-10,
      "Justification": "Brief explanation of any issues identified"
    }
  }
}
\end{verbatim}
\end{small}
- Be objective and thorough in your evaluation, providing clear justifications for your scores.\\
- Remember that you can accept image inputs directly, so you should analyze the images for each criterion.\\
\\
        \end{minipage}
        \end{tabular}
    \end{tcolorbox}
    \vspace{-2mm}
    \caption{The system prompts for obtaining detailed scores from GPT-based evaluators (Part IV).}
    \label{tab:prompt_edit4}
    \end{minipage}
    \vspace{-2mm}
\end{table*}

%% file: supp/tables/win_rate.tex
\begin{table*}[t]
\centering
\caption{Quantitative comparison of model win rates evaluated by GPT-4o and IntJudge. FDT: Force Dividing Tie metric. w/o Tie: Non-Tie case. w/ Tie (0) and w/ Tie Tie(.5): Count a tie as 0 and 0.5 wins for a model in a battle, respectively.}
\label{tab:winrate_supp}
\begin{tabular}{lcccccccc}
\toprule
\multirow{2}{*}{Method} &
\multicolumn{4}{c}{GPT Evaluation} &
\multicolumn{4}{c}{IntJudge Evaluation} \\
\cmidrule(lr){2-5} \cmidrule(lr){6-9}
& FDT & w/o Tie & w/ Tie (0) & w/ Tie (.5) & FDT & w/o Tie & w/ Tie (0) & w/ Tie (.5) \\
\midrule
NExT-GPT                     & 15.70\% & 15.60\% & 15.36\% & 16.13\% & 44.86\% & 32.61\% &  2.57\% & 48.63\% \\
GILL                         & 28.27\% & 28.29\% & 27.75\% & 28.71\% & 34.73\% &  6.52\% &  0.52\% & 46.51\% \\
MiniGPT-5                    & 30.20\% & 29.60\% & 28.84\% & 30.12\% & 31.97\% &  9.52\% &  1.03\% & 45.64\% \\
Show-o                       & 32.55\% & 32.10\% & 31.46\% & 32.46\% & 39.42\% & 13.16\% &  0.90\% & 47.47\% \\
Orthus                       & 34.63\% & 34.20\% & 33.45\% & 34.54\% & 43.05\% & 17.50\% &  1.19\% & 47.80\% \\
SEED-LLAMA                  & 37.86\% & 37.52\% & 36.50\% & 37.86\% & 44.05\% & 17.14\% &  1.02\% & 48.04\% \\
Emu2                         & 41.24\% & 40.87\% & 40.38\% & 40.98\% & 42.69\% & 12.28\% &  1.20\% & 46.30\% \\
Emu3                         & 43.64\% & 43.26\% & 42.44\% & 43.38\% & 49.66\% & 60.00\% &  4.64\% & 50.77\% \\
Anole                        & 49.37\% & 49.18\% & 48.29\% & 49.19\% & 58.38\% & 50.00\% &  2.88\% & 50.00\% \\
VILA-U                       & 49.37\% & 49.45\% & 48.83\% & 49.46\% & 57.01\% & 70.27\% &  4.68\% & 51.35\% \\
SEED-X                       & 49.38\% & 49.00\% & 48.13\% & 49.02\% & 53.65\% & 46.34\% &  3.39\% & 49.73\% \\
Gemini+Flux               & 74.09\% & 75.00\% & 73.58\% & 74.53\% & 58.03\% & 75.00\% &  5.18\% & 51.73\% \\
GPT-4o+DALL-E3               & 83.30\% & 83.51\% & 81.91\% & 82.87\% & \seco{63.46}\% & 86.89\% &  9.27\% & 53.93\% \\
Nano Banana        & \seco{90.99}\% & \seco{91.92}\% & \seco{90.46}\% & \seco{91.25}\% & \best{67.61}\% & \best{94.51}\% & \best{15.22}\% & \best{57.17}\% \\
\rowcolor{Gray}
Wan-Weaver (Ours) & \best{91.84}\% & \best{92.46}\% & \best{91.31}\% & \best{91.93}\% & 62.94\% & \seco{92.86}\% & \seco{11.52}\% & \seco{55.32}\% \\

\bottomrule
\end{tabular}

\end{table*}

%% file: main.bib
@String(CVPR= {IEEE Conf. Comput. Vis. Pattern Recog.})

@String(ICCV= {Int. Conf. Comput. Vis.})

@String(NIPS= {Adv. Neural Inform. Process. Syst.})

@String(ACMMM= {ACM Int. Conf. Multimedia})

@String(ICLR = {Int. Conf. Learn. Represent.})

@String(CVPR  = {CVPR})

@String(ICML  = {ICML})

@String(ICCV  = {ICCV})

@String(NIPS  = {NeurIPS})

@String(ACMMM = {ACM MM})

@String(ICLR  = {ICLR})

@article{bai2025qwen2,
  title={Qwen2. 5-vl technical report},
  author={Bai, Shuai and Chen, Keqin and Liu, Xuejing and Wang, Jialin and Ge, Wenbin and Song, Sibo and Dang, Kai and Wang, Peng and Wang, Shijie and Tang, Jun and others},
  journal={arXiv preprint arXiv:2502.13923},
  year={2025}
}

@article{zhu2025internvl3,
  title={Internvl3: Exploring advanced training and test-time recipes for open-source multimodal models},
  author={Zhu, Jinguo and Wang, Weiyun and Chen, Zhe and Liu, Zhaoyang and Ye, Shenglong and Gu, Lixin and Tian, Hao and Duan, Yuchen and Su, Weijie and Shao, Jie and others},
  journal={arXiv preprint arXiv:2504.10479},
  year={2025}
}

@article{deng2025emerging,
  title={Emerging properties in unified multimodal pretraining},
  author={Deng, Chaorui and Zhu, Deyao and Li, Kunchang and Gou, Chenhui and Li, Feng and Wang, Zeyu and Zhong, Shu and Yu, Weihao and Nie, Xiaonan and Song, Ziang and others},
  journal={arXiv preprint arXiv:2505.14683},
  year={2025}
}

@article{liao2025mogao,
  title={Mogao: An omni foundation model for interleaved multi-modal generation},
  author={Liao, Chao and Liu, Liyang and Wang, Xun and Luo, Zhengxiong and Zhang, Xinyu and Zhao, Wenliang and Wu, Jie and Li, Liang and Tian, Zhi and Huang, Weilin},
  journal={arXiv preprint arXiv:2505.05472},
  year={2025}
}

@article{yang2025mmada,
  title={Mmada: Multimodal large diffusion language models},
  author={Yang, Ling and Tian, Ye and Li, Bowen and Zhang, Xinchen and Shen, Ke and Tong, Yunhai and Wang, Mengdi},
  journal={arXiv preprint arXiv:2505.15809},
  year={2025}
}

@article{wang2025fudoki,
  title={Fudoki: Discrete flow-based unified understanding and generation via kinetic-optimal velocities},
  author={Wang, Jin and Lai, Yao and Li, Aoxue and Zhang, Shifeng and Sun, Jiacheng and Kang, Ning and Wu, Chengyue and Li, Zhenguo and Luo, Ping},
  journal={arXiv preprint arXiv:2505.20147},
  year={2025}
}

@article{kou2024orthus,
  title={Orthus: Autoregressive interleaved image-text generation with modality-specific heads},
  author={Kou, Siqi and Jin, Jiachun and Liu, Zhihong and Liu, Chang and Ma, Ye and Jia, Jian and Chen, Quan and Jiang, Peng and Deng, Zhijie},
  journal={arXiv preprint arXiv:2412.00127},
  year={2024}
}

@inproceedings{wu2024next,
  title={Next-gpt: Any-to-any multimodal llm},
  author={Wu, Shengqiong and Fei, Hao and Qu, Leigang and Ji, Wei and Chua, Tat-Seng},
  booktitle=ICML,
  year={2024}
}

@article{ge2024seed,
  title={Seed-x: Multimodal models with unified multi-granularity comprehension and generation},
  author={Ge, Yuying and Zhao, Sijie and Zhu, Jinguo and Ge, Yixiao and Yi, Kun and Song, Lin and Li, Chen and Ding, Xiaohan and Shan, Ying},
  journal={arXiv preprint arXiv:2404.14396},
  year={2024}
}

@inproceedings{wang2025illume,
  title={Illume: Illuminating your llms to see, draw, and self-enhance},
  author={Wang, Chunwei and Lu, Guansong and Yang, Junwei and Huang, Runhui and Han, Jianhua and Hou, Lu and Zhang, Wei and Xu, Hang},
  booktitle=ICCV,
  year={2025}
}

@inproceedings{tong2025metamorph,
  title={Metamorph: Multimodal understanding and generation via instruction tuning},
  author={Tong, Shengbang and Fan, David and Li, Jiachen and Xiong, Yunyang and Chen, Xinlei and Sinha, Koustuv and Rabbat, Michael and LeCun, Yann and Xie, Saining and Liu, Zhuang},
  booktitle=ICCV,
  year={2025}
}

@article{pan2025transfer,
  title={Transfer between modalities with metaqueries},
  author={Pan, Xichen and Shukla, Satya Narayan and Singh, Aashu and Zhao, Zhuokai and Mishra, Shlok Kumar and Wang, Jialiang and Xu, Zhiyang and Chen, Jiuhai and Li, Kunpeng and Juefei-Xu, Felix and others},
  journal={arXiv preprint arXiv:2504.06256},
  year={2025}
}

@article{ge202seedllama,
  title={Making llama see and draw with seed tokenizer},
  author={Ge, Yuying and Zhao, Sijie and Zeng, Ziyun and Ge, Yixiao and Li, Chen and Wang, Xintao and Shan, Ying},
  journal={arXiv preprint arXiv:2310.01218},
  year={2023}
}

@inproceedings{koh2023gill,
  title={Generating images with multimodal language models},
  author={Koh, Jing Yu and Fried, Daniel and Salakhutdinov, Russ R},
  booktitle=NIPS,
  year={2023}
}

@article{lin2025uniworld,
  title={Uniworld: High-resolution semantic encoders for unified visual understanding and generation},
  author={Lin, Bin and Li, Zongjian and Cheng, Xinhua and Niu, Yuwei and Ye, Yang and He, Xianyi and Yuan, Shenghai and Yu, Wangbo and Wang, Shaodong and Ge, Yunyang and others},
  journal={arXiv preprint arXiv:2506.03147},
  year={2025}
}

@article{li2025uniworld,
  title={Uniworld-V2: Reinforce Image Editing with Diffusion Negative-aware Finetuning and MLLM Implicit Feedback},
  author={Li, Zongjian and Liu, Zheyuan and Zhang, Qihui and Lin, Bin and Yuan, Shenghai and Yan, Zhiyuan and Ye, Yang and Yu, Wangbo and Niu, Yuwei and Yuan, Li},
  journal={arXiv preprint arXiv:2510.16888},
  year={2025}
}

@article{wu2025openuni,
  title={OpenUni: A Simple Baseline for Unified Multimodal Understanding and Generation},
  author={Wu, Size and Wu, Zhonghua and Gong, Zerui and Tao, Qingyi and Jin, Sheng and Li, Qinyue and Li, Wei and Loy, Chen Change},
  journal={arXiv preprint arXiv:2505.23661},
  year={2025}
}

@inproceedings{qu2025tokenflow,
  title={Tokenflow: Unified image tokenizer for multimodal understanding and generation},
  author={Qu, Liao and Zhang, Huichao and Liu, Yiheng and Wang, Xu and Jiang, Yi and Gao, Yiming and Ye, Hu and Du, Daniel K and Yuan, Zehuan and Wu, Xinglong},
  booktitle=CVPR,
  year={2025}
}

@article{huang2025illume+,
  title={Illume+: Illuminating unified mllm with dual visual tokenization and diffusion refinement},
  author={Huang, Runhui and Wang, Chunwei and Yang, Junwei and Lu, Guansong and Yuan, Yunlong and Han, Jianhua and Hou, Lu and Zhang, Wei and Hong, Lanqing and Zhao, Hengshuang and others},
  journal={arXiv preprint arXiv:2504.01934},
  year={2025}
}

@article{team2024chameleon,
  title={Chameleon: Mixed-modal early-fusion foundation models},
  author={Team, Chameleon},
  journal={arXiv preprint arXiv:2405.09818},
  year={2024}
}

@article{chern2024anole,
  title={Anole: An open, autoregressive, native large multimodal models for interleaved image-text generation},
  author={Chern, Ethan and Su, Jiadi and Ma, Yan and Liu, Pengfei},
  journal={arXiv preprint arXiv:2407.06135},
  year={2024}
}

@article{chen2025janus,
  title={Janus-pro: Unified multimodal understanding and generation with data and model scaling},
  author={Chen, Xiaokang and Wu, Zhiyu and Liu, Xingchao and Pan, Zizheng and Liu, Wen and Xie, Zhenda and Yu, Xingkai and Ruan, Chong},
  journal={arXiv preprint arXiv:2501.17811},
  year={2025}
}

@inproceedings{ma2025janusflow,
  title={Janusflow: Harmonizing autoregression and rectified flow for unified multimodal understanding and generation},
  author={Ma, Yiyang and Liu, Xingchao and Chen, Xiaokang and Liu, Wen and Wu, Chengyue and Wu, Zhiyu and Pan, Zizheng and Xie, Zhenda and Zhang, Haowei and Yu, Xingkai and others},
  booktitle=CVPR,
  year={2025}
}

@article{ye2024x,
  title={X-vila: Cross-modality alignment for large language model},
  author={Ye, Hanrong and Huang, De-An and Lu, Yao and Yu, Zhiding and Ping, Wei and Tao, Andrew and Kautz, Jan and Han, Song and Xu, Dan and Molchanov, Pavlo and others},
  journal={arXiv preprint arXiv:2405.19335},
  year={2024}
}

@article{liu2024world,
  title={World model on million-length video and language with blockwise ringattention},
  author={Liu, Hao and Yan, Wilson and Zaharia, Matei and Abbeel, Pieter},
  journal={arXiv preprint arXiv:2402.08268},
  year={2024}
}

@article{zhao2024monoformer,
  title={Monoformer: One transformer for both diffusion and autoregression},
  author={Zhao, Chuyang and Song, Yuxing and Wang, Wenhao and Feng, Haocheng and Ding, Errui and Sun, Yifan and Xiao, Xinyan and Wang, Jingdong},
  journal={arXiv preprint arXiv:2409.16280},
  year={2024}
}

@article{tian2024mm,
  title={Mm-interleaved: Interleaved image-text generative modeling via multi-modal feature synchronizer},
  author={Tian, Changyao and Zhu, Xizhou and Xiong, Yuwen and Wang, Weiyun and Chen, Zhe and Wang, Wenhai and Chen, Yuntao and Lu, Lewei and Lu, Tong and Zhou, Jie and others},
  journal={arXiv preprint arXiv:2401.10208},
  year={2024}
}

@inproceedings{brown2020language,
  title={Language models are few-shot learners},
  author={Brown, Tom and Mann, Benjamin and Ryder, Nick and Subbiah, Melanie and Kaplan, Jared D and Dhariwal, Prafulla and Neelakantan, Arvind and Shyam, Pranav and Sastry, Girish and Askell, Amanda and others},
  booktitle=NIPS,
  year={2020}
}

@article{radford2018improving,
  title={Improving Language Understanding by Generative Pre-Training},
  author={Radford, Alec and Narasimhan, Karthik and Salimans, Tim and Sutskever, Ilya},
  year={2018},
  journal={OpenAI CDN}
}

@article{radford2019language,
  title={Language models are unsupervised multitask learners},
  author={Radford, Alec and Wu, Jeffrey and Child, Rewon and Luan, David and Amodei, Dario and Sutskever, Ilya and others},
  journal={OpenAI blog},
  volume={1},
  number={8},
  pages={9},
  year={2019}
}

@article{li2024llava,
  title={Llava-onevision: Easy visual task transfer},
  author={Li, Bo and Zhang, Yuanhan and Guo, Dong and Zhang, Renrui and Li, Feng and Zhang, Hao and Zhang, Kaichen and Zhang, Peiyuan and Li, Yanwei and Liu, Ziwei and others},
  journal={arXiv preprint arXiv:2408.03326},
  year={2024}
}

@article{zheng2023minigpt,
  title={Minigpt-5: Interleaved vision-and-language generation via generative vokens},
  author={Zheng, Kaizhi and He, Xuehai and Wang, Xin Eric},
  journal={arXiv preprint arXiv:2310.02239},
  year={2023}
}

@inproceedings{alayrac2022flamingo,
  title={Flamingo: a visual language model for few-shot learning},
  author={Alayrac, Jean-Baptiste and Donahue, Jeff and Luc, Pauline and Miech, Antoine and Barr, Iain and Hasson, Yana and Lenc, Karel and Mensch, Arthur and Millican, Katherine and Reynolds, Malcolm and others},
  booktitle=NIPS,
  year={2022}
}

@article{liu2024lumina,
  title={Lumina-mgpt: Illuminate flexible photorealistic text-to-image generation with multimodal generative pretraining},
  author={Liu, Dongyang and Zhao, Shitian and Zhuo, Le and Lin, Weifeng and Xin, Yi and Li, Xinyue and Qin, Qi and Qiao, Yu and Li, Hongsheng and Gao, Peng},
  journal={arXiv preprint arXiv:2408.02657},
  year={2024}
}

@inproceedings{esser2024scaling,
  title={Scaling rectified flow transformers for high-resolution image synthesis},
  author={Esser, Patrick and Kulal, Sumith and Blattmann, Andreas and Entezari, Rahim and M{\"u}ller, Jonas and Saini, Harry and Levi, Yam and Lorenz, Dominik and Sauer, Axel and Boesel, Frederic and others},
  booktitle=ICML,
  year={2024}
}

@article{labs2025flux,
  title={FLUX. 1 Kontext: Flow Matching for In-Context Image Generation and Editing in Latent Space},
  author={Labs, Black Forest and Batifol, Stephen and Blattmann, Andreas and Boesel, Frederic and Consul, Saksham and Diagne, Cyril and Dockhorn, Tim and English, Jack and English, Zion and Esser, Patrick and others},
  journal={arXiv preprint arXiv:2506.15742},
  year={2025}
}

@article{cui2025emu35,
  title={Emu3. 5: Native Multimodal Models are World Learners},
  author={Cui, Yufeng and Chen, Honghao and Deng, Haoge and Huang, Xu and Li, Xinghang and Liu, Jirong and Liu, Yang and Luo, Zhuoyan and Wang, Jinsheng and Wang, Wenxuan and others},
  journal={arXiv preprint arXiv:2510.26583},
  year={2025}
}

@inproceedings{sun2024generative,
  title={Generative multimodal models are in-context learners},
  author={Sun, Quan and Cui, Yufeng and Zhang, Xiaosong and Zhang, Fan and Yu, Qiying and Wang, Yueze and Rao, Yongming and Liu, Jingjing and Huang, Tiejun and Wang, Xinlong},
  booktitle=CVPR,
  year={2024}
}

@article{sun2023emu,
  title={Emu: Generative pretraining in multimodality},
  author={Sun, Quan and Yu, Qiying and Cui, Yufeng and Zhang, Fan and Zhang, Xiaosong and Wang, Yueze and Gao, Hongcheng and Liu, Jingjing and Huang, Tiejun and Wang, Xinlong},
  journal={arXiv preprint arXiv:2307.05222},
  year={2023}
}

@article{wang2024emu3,
  title={Emu3: Next-token prediction is all you need},
  author={Wang, Xinlong and Zhang, Xiaosong and Luo, Zhengxiong and Sun, Quan and Cui, Yufeng and Wang, Jinsheng and Zhang, Fan and Wang, Yueze and Li, Zhen and Yu, Qiying and others},
  journal={arXiv preprint arXiv:2409.18869},
  year={2024}
}

@article{touvron2023llama,
  title={Llama: Open and efficient foundation language models},
  author={Touvron, Hugo and Lavril, Thibaut and Izacard, Gautier and Martinet, Xavier and Lachaux, Marie-Anne and Lacroix, Timoth{\'e}e and Rozi{\`e}re, Baptiste and Goyal, Naman and Hambro, Eric and Azhar, Faisal and others},
  journal={arXiv preprint arXiv:2302.13971},
  year={2023}
}

@inproceedings{chang2022maskgit,
  title={Maskgit: Masked generative image transformer},
  author={Chang, Huiwen and Zhang, Han and Jiang, Lu and Liu, Ce and Freeman, William T},
  booktitle=CVPR,
  year={2022}
}

@inproceedings{li2024autoregressive,
  title={Autoregressive image generation without vector quantization},
  author={Li, Tianhong and Tian, Yonglong and Li, He and Deng, Mingyang and He, Kaiming},
  booktitle=NIPS,
  year={2024}
}

@inproceedings{tian2024visual,
  title={Visual autoregressive modeling: Scalable image generation via next-scale prediction},
  author={Tian, Keyu and Jiang, Yi and Yuan, Zehuan and Peng, Bingyue and Wang, Liwei},
  booktitle=NIPS,
  year={2024}
}

@article{xie2024show,
  title={Show-o: One single transformer to unify multimodal understanding and generation},
  author={Xie, Jinheng and Mao, Weijia and Bai, Zechen and Zhang, David Junhao and Wang, Weihao and Lin, Kevin Qinghong and Gu, Yuchao and Chen, Zhijie and Yang, Zhenheng and Shou, Mike Zheng},
  journal={arXiv preprint arXiv:2408.12528},
  year={2024}
}

@inproceedings{zhoutransfusion,
  title={Transfusion: Predict the Next Token and Diffuse Images with One Multi-Modal Model},
  author={Zhou, Chunting and YU, LILI and Babu, Arun and Tirumala, Kushal and Yasunaga, Michihiro and Shamis, Leonid and Kahn, Jacob and Ma, Xuezhe and Zettlemoyer, Luke and Levy, Omer},
  booktitle=ICLR,
  year={2025}
}

@article{liang2024mixture,
  title={Mixture-of-transformers: A sparse and scalable architecture for multi-modal foundation models},
  author={Liang, Weixin and Yu, Lili and Luo, Liang and Iyer, Srinivasan and Dong, Ning and Zhou, Chunting and Ghosh, Gargi and Lewis, Mike and Yih, Wen-tau and Zettlemoyer, Luke and others},
  journal={arXiv preprint arXiv:2411.04996},
  year={2024}
}

@article{lin2024moma,
  title={Moma: Efficient early-fusion pre-training with mixture of modality-aware experts},
  author={Lin, Xi Victoria and Shrivastava, Akshat and Luo, Liang and Iyer, Srinivasan and Lewis, Mike and Ghosh, Gargi and Zettlemoyer, Luke and Aghajanyan, Armen},
  journal={arXiv preprint arXiv:2407.21770},
  year={2024}
}

@article{yu2023scaling,
  title={Scaling autoregressive multi-modal models: Pretraining and instruction tuning},
  author={Yu, Lili and Shi, Bowen and Pasunuru, Ramakanth and Muller, Benjamin and Golovneva, Olga and Wang, Tianlu and Babu, Arun and Tang, Binh and Karrer, Brian and Sheynin, Shelly and others},
  journal={arXiv preprint arXiv:2309.02591},
  year={2023}
}

@article{wei2025skywork,
  title={Skywork unipic 2.0: Building kontext model with online rl for unified multimodal model},
  author={Wei, Hongyang and Xu, Baixin and Liu, Hongbo and Wu, Cyrus and Liu, Jie and Peng, Yi and Wang, Peiyu and Liu, Zexiang and He, Jingwen and Xietian, Yidan and others},
  journal={arXiv preprint arXiv:2509.04548},
  year={2025}
}

@inproceedings{xiao2025omnigen,
  title={Omnigen: Unified image generation},
  author={Xiao, Shitao and Wang, Yueze and Zhou, Junjie and Yuan, Huaying and Xing, Xingrun and Yan, Ruiran and Li, Chaofan and Wang, Shuting and Huang, Tiejun and Liu, Zheng},
  booktitle=CVPR,
  year={2025}
}

@article{wu2025omnigen2,
  title={OmniGen2: Exploration to Advanced Multimodal Generation},
  author={Wu, Chenyuan and Zheng, Pengfei and Yan, Ruiran and Xiao, Shitao and Luo, Xin and Wang, Yueze and Li, Wanli and Jiang, Xiyan and Liu, Yexin and Zhou, Junjie and others},
  journal={arXiv preprint arXiv:2506.18871},
  year={2025}
}

@article{wang2025ovis,
  title={Ovis-U1 Technical Report},
  author={Wang, Guo-Hua and Zhao, Shanshan and Zhang, Xinjie and Cao, Liangfu and Zhan, Pengxin and Duan, Lunhao and Lu, Shiyin and Fu, Minghao and Chen, Xiaohao and Zhao, Jianshan and others},
  journal={arXiv preprint arXiv:2506.23044},
  year={2025}
}

@article{chen2025blip3,
  title={Blip3-o: A family of fully open unified multimodal models-architecture, training and dataset},
  author={Chen, Jiuhai and Xu, Zhiyang and Pan, Xichen and Hu, Yushi and Qin, Can and Goldstein, Tom and Huang, Lifu and Zhou, Tianyi and Xie, Saining and Savarese, Silvio and others},
  journal={arXiv preprint arXiv:2505.09568},
  year={2025}
}

@article{guo2025m2,
  title={M2-omni: Advancing omni-mllm for comprehensive modality support with competitive performance},
  author={Guo, Qingpei and Song, Kaiyou and Feng, Zipeng and Ma, Ziping and Zhang, Qinglong and Gao, Sirui and Yu, Xuzheng and Sun, Yunxiao and Chang, Tai-Wei and Chen, Jingdong and others},
  journal={arXiv preprint arXiv:2502.18778},
  year={2025}
}

@article{wu2024vila,
  title={Vila-u: a unified foundation model integrating visual understanding and generation},
  author={Wu, Yecheng and Zhang, Zhuoyang and Chen, Junyu and Tang, Haotian and Li, Dacheng and Fang, Yunhao and Zhu, Ligeng and Xie, Enze and Yin, Hongxu and Yi, Li and others},
  journal={arXiv preprint arXiv:2409.04429},
  year={2024}
}

@article{fan2025unifluid,
  title={Unified autoregressive visual generation and understanding with continuous tokens},
  author={Fan, Lijie and Tang, Luming and Qin, Siyang and Li, Tianhong and Yang, Xuan and Qiao, Siyuan and Steiner, Andreas and Sun, Chen and Li, Yuanzhen and Zhu, Tao and others},
  journal={arXiv preprint arXiv:2503.13436},
  year={2025}
}

@article{shi2024lmfusion,
  title={LMFusion: Adapting Pretrained Language Models for Multimodal Generation},
  author={Shi, Weijia and Han, Xiaochuang and Zhou, Chunting and Liang, Weixin and Lin, Xi Victoria and Zettlemoyer, Luke and Yu, Lili},
  journal={arXiv preprint arXiv:2412.15188},
  year={2024}
}

@article{xie2025show,
  title={Show-o2: Improved Native Unified Multimodal Models},
  author={Xie, Jinheng and Yang, Zhenheng and Shou, Mike Zheng},
  journal={arXiv preprint arXiv:2506.15564},
  year={2025}
}

@article{zhang2025nexus,
  title={Nexus-gen: A unified model for image understanding, generation, and editing},
  author={Zhang, Hong and Duan, Zhongjie and Wang, Xingjun and Zhao, Yuze and Lu, Weiyi and Di, Zhipeng and Xu, Yixuan and Chen, Yingda and Zhang, Yu},
  journal={arXiv preprint arXiv:2504.21356},
  year={2025}
}

@article{geng2025x,
  title={X-omni: Reinforcement learning makes discrete autoregressive image generative models great again},
  author={Geng, Zigang and Wang, Yibing and Ma, Yeyao and Li, Chen and Rao, Yongming and Gu, Shuyang and Zhong, Zhao and Lu, Qinglin and Hu, Han and Zhang, Xiaosong and others},
  journal={arXiv preprint arXiv:2507.22058},
  year={2025}
}

@article{liu2024deepseek,
  title={Deepseek-v3 technical report},
  author={Liu, Aixin and Feng, Bei and Xue, Bing and Wang, Bingxuan and Wu, Bochao and Lu, Chengda and Zhao, Chenggang and Deng, Chengqi and Zhang, Chenyu and Ruan, Chong and others},
  journal={arXiv preprint arXiv:2412.19437},
  year={2024}
}

@article{seedream2025seedream,
  title={Seedream 4.0: Toward next-generation multimodal image generation},
  author={Seedream, Team and Chen, Yunpeng and Gao, Yu and Gong, Lixue and Guo, Meng and Guo, Qiushan and Guo, Zhiyao and Hou, Xiaoxia and Huang, Weilin and Huang, Yixuan and others},
  journal={arXiv preprint arXiv:2509.20427},
  year={2025}
}

@article{cao2025hunyuanimage,
  title={Hunyuanimage 3.0 technical report},
  author={Cao, Siyu and Chen, Hangting and Chen, Peng and Cheng, Yiji and Cui, Yutao and Deng, Xinchi and Dong, Ying and Gong, Kipper and Gu, Tianpeng and Gu, Xiusen and others},
  journal={arXiv preprint arXiv:2509.23951},
  year={2025}
}

@article{wan2025wan,
  title={Wan: Open and advanced large-scale video generative models},
  author={Wan, Team and Wang, Ang and Ai, Baole and Wen, Bin and Mao, Chaojie and Xie, Chen-Wei and Chen, Di and Yu, Feiwu and Zhao, Haiming and Yang, Jianxiao and others},
  journal={arXiv preprint arXiv:2503.20314},
  year={2025}
}

@inproceedings{liu2023visual,
  title={Visual instruction tuning},
  author={Liu, Haotian and Li, Chunyuan and Wu, Qingyang and Lee, Yong Jae},
  booktitle=NIPS,
  year={2023}
}

@misc{Imagen4,
  title         = {Imagen4},
  author = {Google},
  howpublished  = {\emph{Imagen4}. \url{https://deepmind.google/models/imagen/}},
  url           = "https://deepmind.google/models/imagen/",
  year = {2025}
}

@misc{Gemini2.5,
  title         = {Gemini2.5},
  author = {Google Deepmind},
  howpublished  = {\emph{Gemini2.5}. \url{https://gemini.google.com/}},
  url           = "https://gemini.google.com/",
  year={2025}
}

@article{byun2024ares,
  title={ARES: Alternating reinforcement learning and supervised fine-tuning for enhanced multi-modal chain-of-thought reasoning through diverse AI feedback},
  author={Byun, Ju-Seung and Chun, Jiyun and Kil, Jihyung and Perrault, Andrew},
  journal={arXiv preprint arXiv:2407.00087},
  year={2024}
}

@inproceedings{mu2023embodiedgpt,
  title={Embodiedgpt: Vision-language pre-training via embodied chain of thought},
  author={Mu, Yao and Zhang, Qinglong and Hu, Mengkang and Wang, Wenhai and Ding, Mingyu and Jin, Jun and Wang, Bin and Dai, Jifeng and Qiao, Yu and Luo, Ping},
  booktitle=NIPS,
  year={2023}
}

@article{claman2024artificial,
  title={Artificial intelligence in dental education: opportunities and challenges of large Language models and multimodal foundation models},
  author={Claman, Daniel and Sezgin, Emre and others},
  journal={JMIR medical education},
  volume={10},
  number={1},
  pages={e52346},
  year={2024},
  publisher={JMIR Publications Inc., Toronto, Canada}
}

@article{latif2023artificial,
  title={Artificial general intelligence (AGI) for education},
  author={Latif, Ehsan and Mai, Gengchen and Nyaaba, Matthew and Wu, Xuansheng and Liu, Ninghao and Lu, Guoyu and Li, Sheng and Liu, Tianming and Zhai, Xiaoming},
  journal={arXiv preprint arXiv:2304.12479},
  volume={1},
  pages={1--34},
  year={2023},
  publisher={May}
}

@inproceedings{ko2023large,
  title={Large-scale text-to-image generation models for visual artists’ creative works},
  author={Ko, Hyung-Kwon and Park, Gwanmo and Jeon, Hyeon and Jo, Jaemin and Kim, Juho and Seo, Jinwook},
  booktitle={Proceedings of the 28th international conference on intelligent user interfaces},
  pages={919--933},
  year={2023}
}

@article{stella2023can,
  title={How can LLMs transform the robotic design process?},
  author={Stella, Francesco and Della Santina, Cosimo and Hughes, Josie},
  journal={Nature machine intelligence},
  volume={5},
  number={6},
  pages={561--564},
  year={2023},
  publisher={Nature Publishing Group UK London}
}

@inproceedings{zhou2025opening,
  title={OpenING: A Comprehensive Benchmark for Judging Open-ended Interleaved Image-Text Generation},
  author={Zhou, Pengfei and Peng, Xiaopeng and Song, Jiajun and Li, Chuanhao and Xu, Zhaopan and Yang, Yue and Guo, Ziyao and Zhang, Hao and Lin, Yuqi and He, Yefei and others},
  booktitle=CVPR,
  year={2025}
}

@inproceedings{an2024openleaf,
  title={Openleaf: A novel benchmark for open-domain interleaved image-text generation},
  author={An, Jie and Yang, Zhengyuan and Li, Linjie and Wang, Jianfeng and Lin, Kevin and Liu, Zicheng and Wang, Lijuan and Luo, Jiebo},
  booktitle=ACMMM,
  year={2024}
}

@article{liu2024holistic,
  title={Holistic evaluation for interleaved text-and-image generation},
  author={Liu, Minqian and Xu, Zhiyang and Lin, Zihao and Ashby, Trevor and Rimchala, Joy and Zhang, Jiaxin and Huang, Lifu},
  journal={arXiv preprint arXiv:2406.14643},
  year={2024}
}

@article{chen2024interleaved,
  title={Interleaved scene graphs for interleaved text-and-image generation assessment},
  author={Chen, Dongping and Chen, Ruoxi and Pu, Shu and Liu, Zhaoyi and Wu, Yanru and Chen, Caixi and Liu, Benlin and Huang, Yue and Wan, Yao and Zhou, Pan and others},
  journal={arXiv preprint arXiv:2411.17188},
  year={2024}
}

@inproceedings{zhai2023sigmoid,
  title={Sigmoid loss for language image pre-training},
  author={Zhai, Xiaohua and Mustafa, Basil and Kolesnikov, Alexander and Beyer, Lucas},
  booktitle=ICCV,
  year={2023}
}

@inproceedings{lipmanflow,
  title={Flow Matching for Generative Modeling},
  author={Lipman, Yaron and Chen, Ricky TQ and Ben-Hamu, Heli and Nickel, Maximilian and Le, Matthew},
  booktitle=ICLR,
  year={2023}
}

@misc{openai2025gpt4o,
  author       = {OpenAI},
  title        = {Hello {GPT-4o}},
  year         = {2025},
  howpublished = {\url{https://openai.com/index/hello-gpt-4o/}},
  note         = {Accessed: 2025-10-11}
}

@misc{google2025nano,
  author       = {Google},
  title        = {Gemini-2.5-Image},
  year         = {2025},
  howpublished = {https://gemini.google/overview/image-generation/}}

@article{betker2023improving,
  title={Improving image generation with better captions},
  author={Betker, James and Goh, Gabriel and Jing, Li and Brooks, Tim and Wang, Jianfeng and Li, Linjie and Ouyang, Long and Zhuang, Juntang and Lee, Joyce and Guo, Yufei and others},
  journal={Computer Science},
  volume={2},
  number={3},
  pages={8},
  year={2023}
}

@article{team2023gemini,
  title={Gemini: a family of highly capable multimodal models},
  author={Team, Gemini and Anil, Rohan and Borgeaud, Sebastian and Wu, Yonghui and Alayrac, Jean-Baptiste and Yu, Jiahui and Soricut, Radu and Schalkwyk, Johan and Dai, Andrew M and Hauth, Anja and others},
  journal={arXiv preprint arXiv:2312.11805},
  year={2023}
}

@article{lu2025ovis2,
  title={Ovis2. 5 technical report},
  author={Lu, Shiyin and Li, Yang and Xia, Yu and Hu, Yuwei and Zhao, Shanshan and Ma, Yanqing and Wei, Zhichao and Li, Yinglun and Duan, Lunhao and Zhao, Jianshan and others},
  journal={arXiv preprint arXiv:2508.11737},
  year={2025}
}

@article{liu2025step1x,
  title={Step1x-edit: A practical framework for general image editing},
  author={Liu, Shiyu and Han, Yucheng and Xing, Peng and Yin, Fukun and Wang, Rui and Cheng, Wei and Liao, Jiaqi and Wang, Yingming and Fu, Honghao and Han, Chunrui and others},
  journal={arXiv preprint arXiv:2504.17761},
  year={2025}
}

@misc{liu2024llavanext,
    title={LLaVA-NeXT: Improved reasoning, OCR, and world knowledge},
    url={https://llava-vl.github.io/blog/2024-01-30-llava-next/},
    author={Liu, Haotian and Li, Chunyuan and Li, Yuheng and Li, Bo and Zhang, Yuanhan and Shen, Sheng and Lee, Yong Jae},
    month={January},
    year={2024}
}

@article{podell2023sdxl,
  title={Sdxl: Improving latent diffusion models for high-resolution image synthesis},
  author={Podell, Dustin and English, Zion and Lacey, Kyle and Blattmann, Andreas and Dockhorn, Tim and M{\"u}ller, Jonas and Penna, Joe and Rombach, Robin},
  journal={arXiv preprint arXiv:2307.01952},
  year={2023}
}

@article{wang2025skywork,
  title={Skywork unipic: Unified autoregressive modeling for visual understanding and generation},
  author={Wang, Peiyu and Peng, Yi and Gan, Yimeng and Hu, Liang and Xie, Tianyidan and Wang, Xiaokun and Wei, Yichen and Tang, Chuanxin and Zhu, Bo and Li, Changshi and others},
  journal={arXiv preprint arXiv:2508.03320},
  year={2025}
}

@inproceedings{brooks2023instructpix2pix,
  title={Instructpix2pix: Learning to follow image editing instructions},
  author={Brooks, Tim and Holynski, Aleksander and Efros, Alexei A},
  booktitle=CVPR,
  year={2023}
}

@inproceedings{zhang2023magicbrush,
  title={Magicbrush: A manually annotated dataset for instruction-guided image editing},
  author={Zhang, Kai and Mo, Lingbo and Chen, Wenhu and Sun, Huan and Su, Yu},
  booktitle=NIPS,
  year={2023}
}

@inproceedings{yu2025anyedit,
  title={Anyedit: Mastering unified high-quality image editing for any idea},
  author={Yu, Qifan and Chow, Wei and Yue, Zhongqi and Pan, Kaihang and Wu, Yang and Wan, Xiaoyang and Li, Juncheng and Tang, Siliang and Zhang, Hanwang and Zhuang, Yueting},
  booktitle=CVPR,
  year={2025}
}

@article{zhang2025context,
  title={In-context edit: Enabling instructional image editing with in-context generation in large scale diffusion transformer},
  author={Zhang, Zechuan and Xie, Ji and Lu, Yu and Yang, Zongxin and Yang, Yi},
  journal={arXiv preprint arXiv:2504.20690},
  year={2025}
}
